%% file: top.tex
\documentclass[10pt,twocolumn,letterpaper]{article}

\usepackage{cvpr}

\usepackage{graphicx}
\usepackage{amsmath,amssymb} 
\usepackage{color}
\usepackage{times}
\usepackage{epsfig}
\usepackage{multirow}
\usepackage{subcaption}
\usepackage{verbatim}
\usepackage{gensymb}
\usepackage{multirow}
\usepackage[toc,page]{appendix}
\usepackage{algorithm}
\usepackage[noend]{algpseudocode}
\usepackage{hyperref}

\begin{document}
\cvprfinalcopy

\input{notation}

\title{Find your Way by Observing the Sun and Other Semantic Cues} 

\author{Wei-Chiu Ma$^{1}$ \; Shenlong Wang$^2$ \; Marcus A. Brubaker$^2$ \; Sanja Fidler$^2$ \; Raquel Urtasun$^2$ \\ \\
$^1$Carnegie Mellon University \quad $^2$University of Toronto
}

\maketitle

\vspace{-1mm}
\input{abstract}
\vspace{-2mm}
\input{intro}

\input{related}
\input{sun}

\input{method}

\input{results}

\input{conc}


{\small
\bibliographystyle{ieee}
\bibliography{egbib}
}


\end{document}

%% file: notation.tex
\newcommand{\myincframe}[1]{\includegraphics[width=3.2cm,trim=0.05cm 0.05cm 0cm 0cm,clip]{#1.pdf}}
\newcommand{\myincframesupp}[1]{\includegraphics[width=3.2cm,trim=0.05cm 0.05cm 0cm 0cm,clip]{#1.pdf}}
\newcommand{\myincframeamb}[1]{\includegraphics[width=1.9cm,trim=0.05cm 0.05cm 0cm 0cm,clip]{#1.pdf}}
\newcommand{\myincframefull}[1]{\includegraphics[width=4.1cm,trim=0.35cm 0.35cm 0.15cm 0.15cm,clip]{#1.pdf}}

\newcommand{\marcus}[1]{\textcolor{red}{\textbf{[Marcus: #1]}}}

\newcommand{\mvec}[1]{ {\mathbf{#1}} }
\newcommand{\mmat}[1]{ {\mathbf{#1}} }

\newcommand{\pos}{ {\mvec{p}} }
\newcommand{\state}{ {\mathbf{x}} }
\newcommand{\obs}{ {\mvec{y}} }
\newcommand{\streetState}{ {\mvec{s}} }

\newcommand{\transAng}{{\theta}}
\newcommand{\streetAngAR}{{\gamma}}

\newcommand{\streetNode}{u}
\newcommand{\streetDist}{d}
\newcommand{\streetAng}{\theta}
\newcommand{\streetArcC}{ {\mvec{c}} }
\newcommand{\streetArcR}{r}
\newcommand{\streetArcAng}{\psi}
\newcommand{\streetLen}{\ell}
\newcommand{\streetCurvature}{\alpha}
\newcommand{\streetOrient}{\beta}
\newcommand{\streetRoadType}{R}
\newcommand{\streetInterType}{I}
\newcommand{\streetSpeedLimit}{V}

\newcommand{\figref}[1]{Fig.~\ref{#1}}
\newcommand{\equref}[1]{Eq.~\eqref{#1}}
\newcommand{\secref}[1]{Sec.~\ref{#1}}
\newcommand{\tabref}[1]{Tab.~\ref{#1}}
\newcommand{\thmref}[1]{Theorem~\ref{#1}}
\newcommand{\prgref}[1]{Program~\ref{#1}}
\renewcommand{\algref}[1]{Alg.~\ref{#1}}
\newcommand{\clmref}[1]{Claim~\ref{#1}}
\newcommand{\lemref}[1]{Lemma~\ref{#1}}
\newcommand{\ptyref}[1]{Property~\ref{#1}}
\newcommand{\propref}[1]{Proposition~\ref{#1}}

\newcommand{\raquel}[1]{{\color{red}Raquel: {#1}}}
\newcommand{\weichiu}[1]{{\color{red}Wei-Chiu: {#1}}}
\newcommand{\shenlong}[1]{{\color{blue}Shenlong: {#1}}}

\newcommand{\todo}[1]{{\color{red} \textbf{[Todo: #1]}}}
\newcommand{\ignoretext}[1]{}

\def\eg{\emph{e.g}.} \def\Eg{\emph{E.g}.}
\def\ie{\emph{i.e}.} \def\Ie{\emph{I.e}.}
\def\cf{\emph{c.f}.} \def\Cf{\emph{C.f}.}
\def\etc{\emph{etc}.} \def\vs{\emph{vs}.}
\def\wrt{w.r.t.} \def\dof{d.o.f.}
\def\etal{\emph{et al}.}

%% file: abstract.tex
 \vspace{-0.5cm}
\begin{abstract}
In this paper we present a robust, efficient and affordable approach to self-localization which does not require neither GPS nor knowledge about the appearance of the world. Towards this goal, we utilize freely available cartographic maps and derive a probabilistic model that exploits
semantic cues in the form of sun direction, presence of an intersection, road type,  speed limit as well as the ego-car trajectory in order to produce very reliable localization results.  Our experimental evaluation shows that our approach can localize much faster (in terms of driving time) with less computation and more robustly  than competing approaches, which ignore semantic information. 
\end{abstract}

%% file: intro.tex
 
\section{Introduction}

Self-localization is a crucial component required to make  self-driving cars a reality. 
An autonomous system has to be able to drive from point A to point B, park itself and recharge its battery when needed. With the availability of maps, semantic scene understanding  becomes easier when the car is localized as strong priors from the map can be exploited~\cite{WangCVPR15,nyc3dcar}. Self-localization is also key for map building.

The most commonly used self-localization technique is the Global Positioning System (GPS), which exploits triangulation from different satellites to determine the position of the GPS device. However,  low-cost  GPS systems are not reliable enough for applications such as robotics or self-driving cars.   The presence of skyscrapers, signal jammers and narrow streets are common sources of problems, that make these systems non robust. 

To overcome the limitations of GPS, many {\it place recognition} techniques have been developed in the past few years. These approaches record how the ``world" looks like either in terms of geometry (e.g., LIDAR point clouds) or visual features, and frame  localization as  a retrieval task, where one has to  search for a similar place in the large-scale dataset of the world~\cite{Baatz12,Dewri13,Hays08,Li12,lin2015learning,Sattler11,Schindler07,Zhang06}. This is typically combined with GPS, which narrows down the region of interest where the search needs to be performed. 
The main limitation of place-recognition approaches is that they require an up-to-date representation of the whole world. This is far from trivial, as the world is constantly changing and in the case of visual features, one needs to capture over different seasons, weather conditions and possibly times of the day.
Privacy is also an issue, as recording is currently illegal in countries such as Germany, where cameras can be used for driving but their content cannot be stored.

With these problems in mind, Brubaker et al.~\cite{brubaker2013lost} developed an approach to self-localization that exploits freely available maps from  OpenStreetMaps (OSM)   and  localizes based solely on visual odometry. The idea behind is that the vehicle's trajectory is a very strong indicator of which roads the vehicle could potentially be driving on, and if one drives long enough, the car's  possible location can be narrowed down to a single mode in the map. 
This paradigm is much more appealing than place recognition approaches, as it does not require to know/store the appearance of the world,  and only a cartographic map of the road topology is necessary. 
 
Brubaker et al. \cite{brubaker2013lost} showed very impressive results  in terms of localization accuracy, reaching the precision  of the map. However, their approach suffers from three main problems. First, it can fail to localize  in very dense road maps as it relies solely on the uniqueness of the ego-motion. 
Second, the time to localization remains fairly large, reducing its applicability. Last, the computational complexity is a function of the uncertainty in the map, which remains fairly large when dealing with maps that have repetitive structures (e.g., Manhattan grid). 

\begin{figure*}[t]
\vspace{-0.7cm}
\centering
\includegraphics[width=0.9\linewidth]{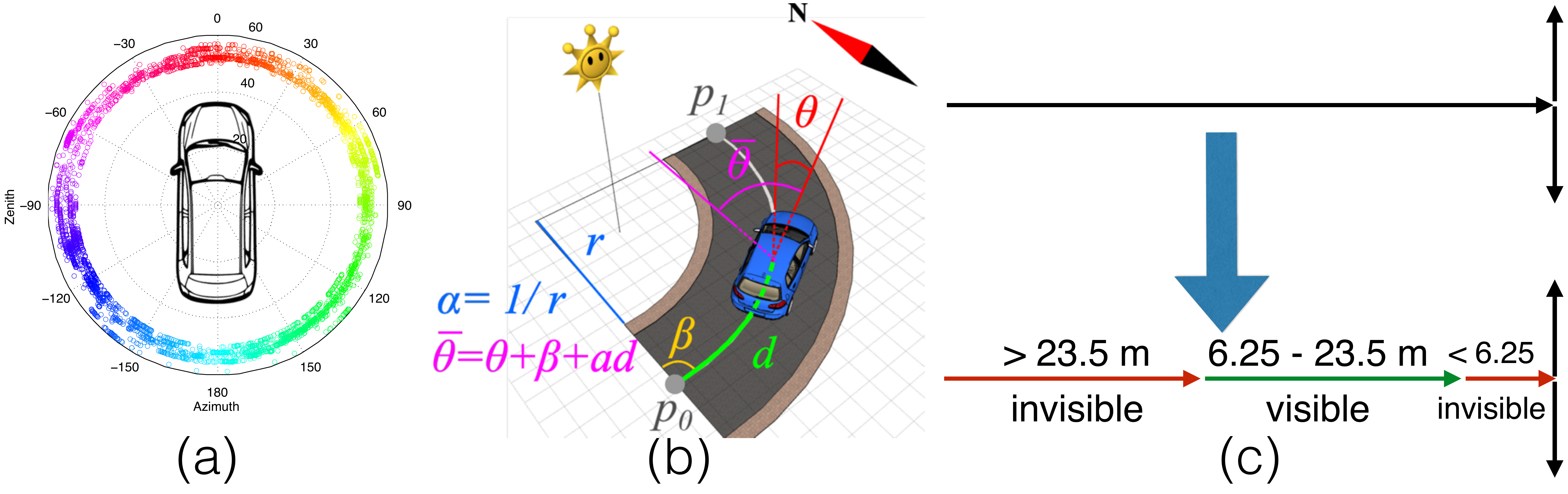}
\vspace{-0.4cm}
\caption{(a) The sun distribution in the KITTI Raw Dataset. (b) The parameterization used to describe a street  segment as well as the position and the heading of a vehicle. (c) To encode intersection information, we partition road segments based on the visibility of an intersection.}
\label{fig:merge-three}
\vspace{-0.3cm}
\end{figure*}

In this paper, we push the limit of vision-based navigation systems. We propose to exploit semantics to reduce localization time and the computational cost, as well as to increase the number of sequences that can be successfully localized.   
We remain in the scenario where  the appearance of the world is unknown in order to create affordable solutions to self-localization.  
Towards this goal, we develop a novel probabilistic localization approach which makes use of 
four different types of semantics in addition to the vehicle's trajectory. The most important cue we exploit is the sun, which (if we know the time of the day) can act as a compass, providing  information about the absolute orientation of the vehicle.  
This provides  very complementary information to visual odometry, which is rotation invariant as we do not know the initial orientation of the car. Additionally, we  exploit  the presence/absence of an intersection, the type of road that we are driving on, as well as the roads' speed limit. 
These cues can help us   narrow down  the set of  possible car's locations in the map in a very short time,  reducing the computational demand of our  localization algorithm.

Estimating the sun direction is not an easy task, as the sun might not be present in the image. Traditional approaches~\cite{lalonde2009estimating} detect shadows and estimate the sun direction from them. Unfortunately,  estimating  shadows is far from trivial, and as a consequence  existing approaches fail  to provide accurate estimates of  the sun direction.
Instead, we take an alternative approach and employ a convolutional network to directly estimate the sun direction from a single image. 
As shown in our experiments, this simple approach works remarkably well in real-world driving scenarios. Interestingly, 
conv3 and conv4 activations fire at both shading and shadow regions. 
We also employ deep learning to estimate the presence/absence of an intersection and the road type. Importantly, we show that  one can use OSM and the time of the day to create automatic labels for all the  tasks (i.e., sun estimation, road type, presence/absence of an intersection).

We demonstrate the effectiveness of our approach on the challenging KITTI dataset~\cite{Geiger2012CVPR} and show that we can localize much faster and with a lower computational cost than~\cite{brubaker2013lost}. Furthermore, we successfully localize in scenarios where \cite{brubaker2013lost} fails. 
Next, we  discuss related work, how to use deep learning to extract semantics as well as  our novel  self-localization approach. We then evaluate our approach and conclude.

%% file: related.tex
 \vspace{-2mm}
\section{Related Work}
\vspace{-1mm}

Localization in maps has been long studied in robotics, typically
with particle-based Monte Carlo methods~\cite{Dellaert99,Fox99,Gutmann98,Oh04}.
These approaches mainly employ  wheel odometry or
depth measurements as observations.
Maps have also been used to improve upon noisy GPS signals~\cite{Bonnabel2011,ElNajjar2005,Fouque2012,Guivant2007}. In contrast, 
in this  work we assume  no knowledge of the initial position of the vehicle 
beyond a broad  region of interest (which contains more than 2000km of road).

Place recognition methods  attempt to perform self-localization without GPS,  by searching for similar scenes in a large database of geo-tagged images ~\cite{Baatz12,Hays08,lin2015learning,Sattler11,Zhang06,Li06,workman2015wide}, 
3D point clouds \cite{Li12,moosmann2013joint,wolcott2014visual,Bansal14}, 3D line segments \cite{Bansal14} or driving trajectories \cite{Dewri13}.  
These approaches typically have limitations as the database of
images must be captured and kept up-to-date. Very recent work~\cite{nelson2015dusk,linegar2015work} has started to make image-based localization invariant to certain appearance changes, however, they are still not very robust to severe changes in visual appearance due to weather and illumination.

Localization in road maps using visual cues in the form of the car's ego-trajectory was recently 
studied in~\cite{brubaker2013lost,floros13}. The advantage of this line of work over the retrieval-based approaches is that it only requires  a cartographic map of the road network. Projects such as OpenStreetMap already provide world coverage and can be freely downloaded from the web. 
While~\cite{floros13} requires an initial estimate of position and was only tested in very small maps,~\cite{brubaker2013lost} showed impressive localization results where the region of interest was the whole city. 
\cite{brubaker2013lost} relies solely on visual odometry, ignoring other
visual and semantic cues which could be exploited for the localization task. In contrast, in this paper we propose to use the sun direction, the presence/absence of an intersection, the type of road we are driving on as well as the speed limit in addition to visual odometry as cues for localization. As shown in our experiments, we are able to localize faster, with less computation and with a higher success rate than~\cite{brubaker2013lost}. 

Recent work exploited semantics for geo-localization. Shadows,  direction of the sun~\cite{JunejoECCV2008,WuCVIU2010} and even rainbows~\cite{WorkmanECCV2014} have been used for very rough localization (roughly 100km error) and camera 
calibration. These approaches require a long video recorded with a stationary camera, which is not realistic in autonomous driving scenarios. In~\cite{LambertJFR2012}, visual odometry was combined with sun and gravity sensors. However, additional sensors are required for this method, while our approach directly reasons about the sun direction from images. Another interesting work exploited semantic labeling from a single image and matching with the GIS dataset~\cite{castaldo_vss15}. However, using semantics from a single image alone cannot achieve a meter-level accuracy in large-scale urban environments, where the semantic layout of the scene is very repetitive.


Semantic cues imposed by maps have also been used for indoor settings such as localization in apartments~\cite{ApartmentsCVPR15} and museums~\cite{Brualla14}. Their cues are tailored to static imagery, and thus not directly relevant for our scenario.

\begin{figure}[t]
\vspace{-0.4cm}
\centering
\includegraphics[width=0.98\linewidth]{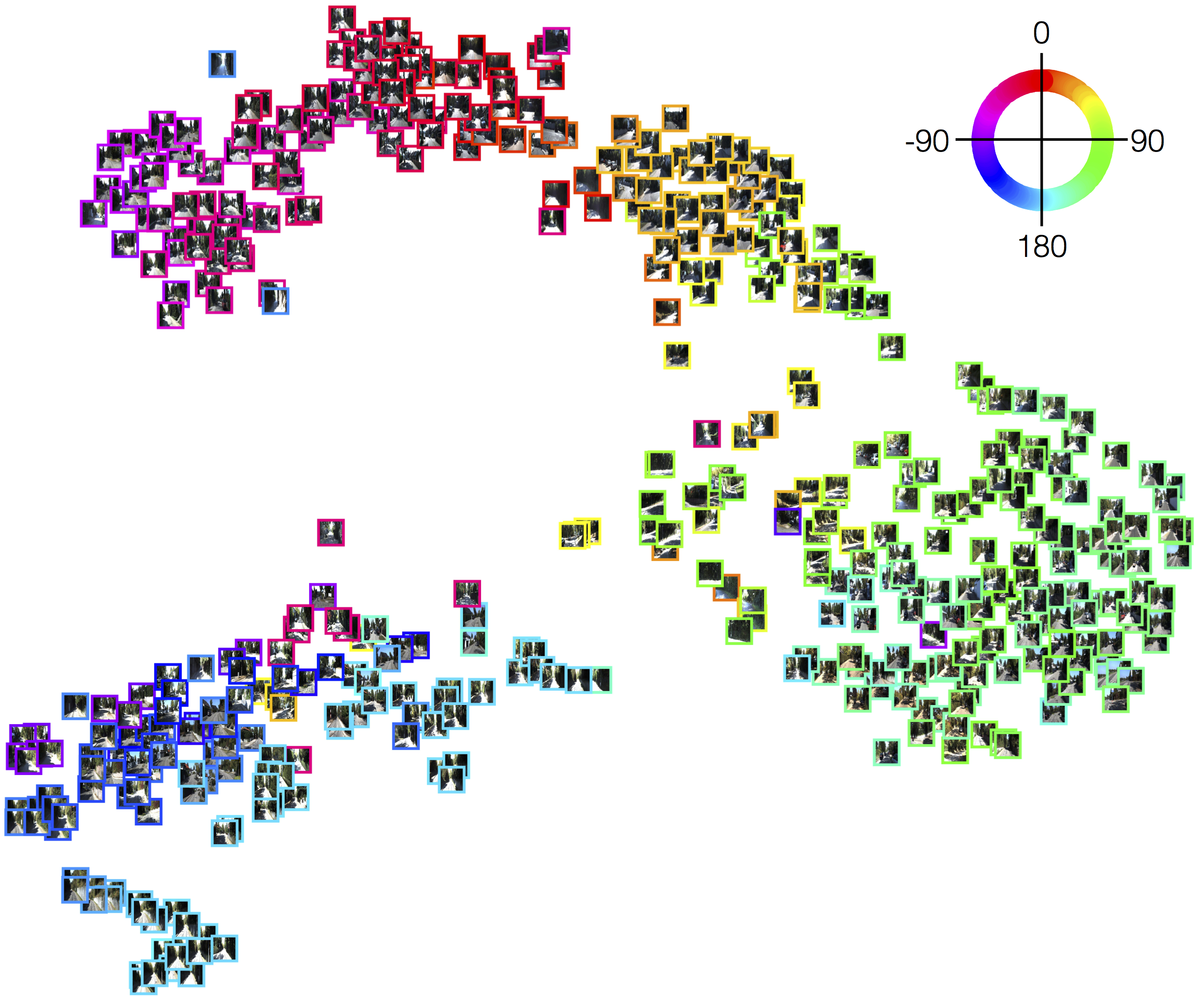}
\vspace{-0.2cm}
\caption{2D embedding of the Sun-CNN feature space with t-SNE. Color of the image border denotes  ground truth relative sun position, \ie\ {\color{red} \bf red}  sun is in front of the vehicle, {\color[rgb]{0,0.7,0} \bf green}  sun is on the right, {\color{cyan} \bf cyan}  sun is on the back, and {\color{magenta} \bf magenta}  sun is on the left. Sun-CNN not only effectively separates images showing different sun directions, but also preserves the \emph{relative relationship}, \eg\ images where sun are on the {\color{red} \bf front} lie between images where sun are on the {\color{magenta} \bf left front} and {\color[rgb]{0.9,0.8,0} \bf right front}}
\label{fig:tSNEsun}
\vspace{-7mm}
\end{figure}

\begin{figure}[t]
\vspace{-0.4cm}
\centering
\includegraphics[width=0.98\linewidth]{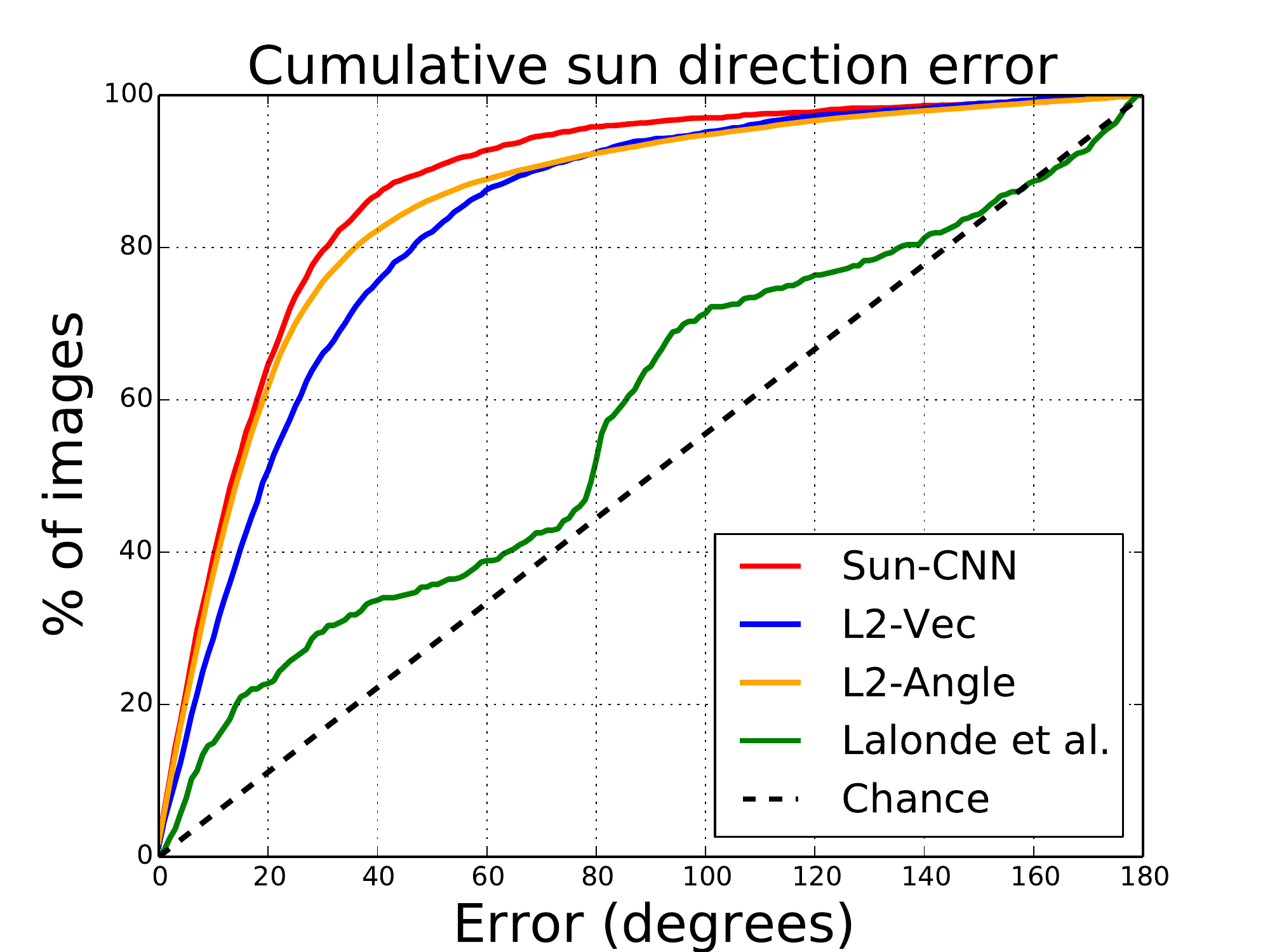}
\vspace{-0.2cm}
\caption{Cumulative sun direction error (angle between prediction and ground truth direction) for different methods on \emph{KITTI-Sun} dataset. Our Sun-CNN outperforms previous state-of-the-art \cite{lalonde2009estimating} and other parameterizations by a large margin.}
\label{fig:sun-curve}
\vspace{-7mm}
\end{figure}

%% file: sun.tex
\section{Deep Learning for Semantics}
In this section we describe how we employ deep learning to estimate  sun direction, road type as well as the presence/absence of an intersection. 

\subsection{Sun Direction Estimation}
\label{sec:sun-direction-estimation}

We start our discussion by  describing how to estimate relative sun direction from a single image. Given the time of the day and a coarse geo-location, we can recover the absolute camera pose (\ie, vehicle's heading direction) by estimating the sun direction in the image, which can be used as a  compass. As shown in the next section knowing the sun direction will significantly reduce localization time and computational cost. 

Estimating illumination from a \textit{single} RGB image is, however,  an ill-posed problem due to the unknown depth/texture/albedo of the scene. Yet humans are good at estimating the light direction  with the help of some visual cues \eg, shadows, shading, over-exposed regions in the sky. For instance, if the buildings on the left side of the image are much brighter than the ones on the right side, we can infer that the sun is most likely  on the right hand side.  If we had an effective shadow detector and we knew the geometry of the scene, we could  estimate  the sun position. Unfortunately,  neither shadow detectors nor shading estimation algorithms are  good enough. Instead, in this paper we adopt a convolutional neural network to automatically learn and capture all kinds of visual cues that may help to estimate relative sun position.

\begin{figure*}[t]
\vspace{-0.3cm}
\centering
\includegraphics[width=0.98\linewidth]{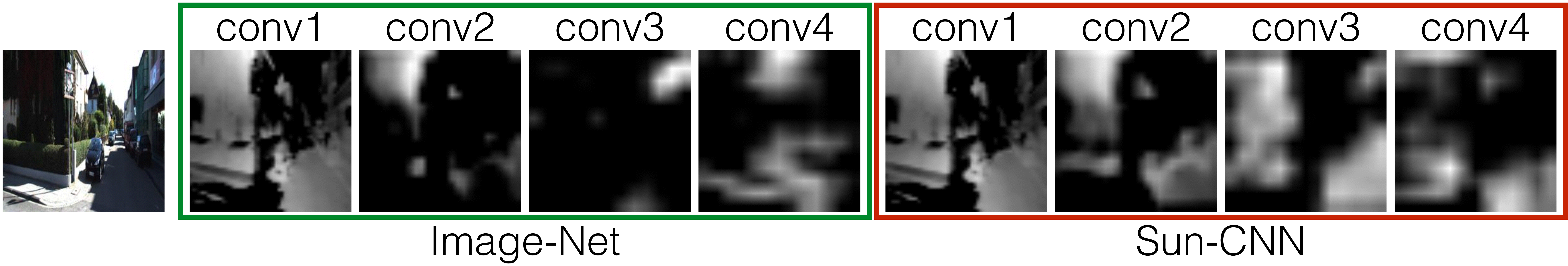}
\vspace{-0.2cm}
\caption{{\bf Comparison between ImageNet-CNN and Sun-CNN: }the activation map of units at different layers for an input image. As the network goes deeper, the original ImageNet-CNN starts to capture certain high-level concepts, while our Sun-CNN focus on detecting the local illumination variation which is of crucial importance in detecting shadow and inferring sun position.}
\label{fig:imagenet-vs-sun-cnn}
\vspace{-2.5mm}
\end{figure*}

\begin{figure*}[t]
\vspace{-0.2cm}
\centering
\includegraphics[width=0.98\linewidth]{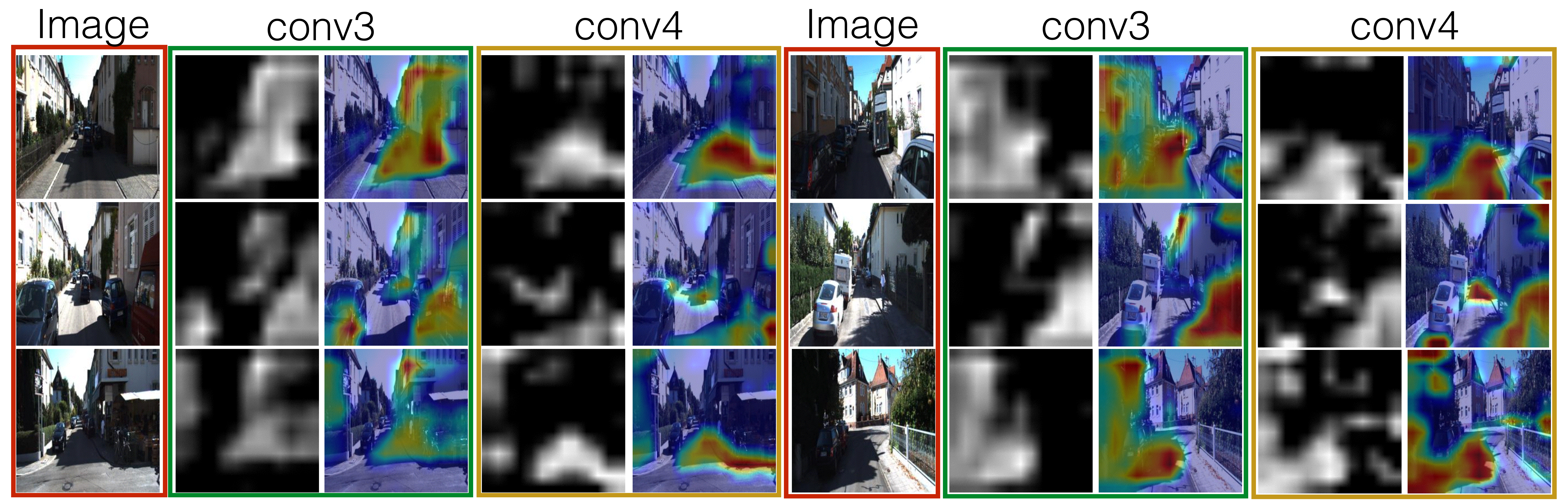}
\vspace{-0.2cm}
\caption{{\bf Shading/shadow detectors emerge in Sun-CNN: }Test images and the corresponding activation maps of certain units in conv3 and conv4 layers of Sun-CNN. Despite being trained on \emph{image-level} label (the relative sun position), our Sun-CNN automatically learns to fire on shadings (conv3) and shadows (conv4).}
\label{fig:sun-activation}
\vspace{-2.5mm}
\end{figure*}

In order to perform end-to-end training of the neural net we must have enough labeled data. However, to our knowledge no dataset with ground-truth labeled sun direction is large enough to train a deep neural network. Fortunately, given the timestamp and a coarse location (which we infer from GPS+IMU on KITTI \cite{Geiger2012CVPR}), we can calculate the absolute sun position in the sky under a horizontal coordinate system with the solar positioning algorithm \cite{solar}. We refer the readers to the appendix\footnote{\href{http://www.cs.toronto.edu/~weichium/geo/appendix.pdf}{http://www.cs.toronto.edu/{\textasciitilde}weichium/geo/appendix.pdf}} for a detailed explanation of this algorithm. We thus automatically generate  labels from the KITTI Raw dataset \cite{Geiger2012CVPR} to form our {\it KITTI-Sun} dataset. As adjacent frames are visually very similar, we subsample the videos at  1 frame/s, resulting in $3314$ images. Note that the distribution of sun directions is quite uniform for our  {\it KITTI-Sun}  dataset as depicted in  \figref{fig:merge-three}(a). 
 
To ensure that our network takes into account the geodesics of the rotation manifold, we parameterize the sun direction with a two dimension unit vector. Our network structure is adapted from AlexNet~\cite{krizhevsky2012imagenet}, where we replace the softmax layer with two continuous output variables representing the predicted 2D vector. By minimizing the distance between the ground truth vector and the output, the network is learning how to predict the sun direction correctly. We adopt cosine distance as our distance metric which measures the angle between the two vectors and in practice performs the best (see \secref{sec:exp-deep} for more details).

\subsection{Intersection Classification}
\label{sec:intersection-classification}
Unlike the sun direction, the presence of an intersection in an image cannot be directly estimated from  GPS+IMU. An alternative is to use  crowd-sourcing systems such as Amazon Mechanical Turk (MTurk). Labeling images, however, is an expensive process as  a quality control process is frequently required in order to sanitize the annotations. Instead, we exploit  GPS/IMU  as well as map data (i.e., OSM) to automatically generate ground truth labels.

Similar to the {\it KITTI-Sun} dataset, we subsample the KITTI-Raw dataset~\cite{Geiger2012CVPR} at 1 frame/s.  We then locate the camera's position and estimate the visible area in the map using the GPS+IMU information. We further exploit the fact that the field-of-view of  KITTI is 135 degrees, and defined the intersection visible area to be a sector with radius  (6.25m-23m). The radius is selected via an empirical in-house user study according to whether humans can reliably determine the presence of
intersection. We then automatically label each image as containing an intersection if there is one  in the sector of interest of the OSM map. Using this  procedure, we obtained the {\it KITTI-Intersection} dataset, consisting of  $3314$ images,  $518$ of which contain an intersection. Note that although we focused on images from KITTI, our automatic labeling procedure can be generalized to other  geo-tagged images.
Our intersection classification network is adapted from GoogleLeNet~\cite{szegedy2014going}.

\subsection{Road Type Classification}
\label{sec:road-type-classification}
We  subsample the KITTI-Raw dataset~\cite{Geiger2012CVPR} at 1 frame/s, and   project the camera location of each image using GPS+IMU onto the nearest street segment on the OSM map. We then use the   road type category provided by OSM to automatically create  labels for the road-type classification task. 
We collapse labels \{{\it trunk, trunk-link, motorway, motorway-link}\},  into a single type  {\it highway}, and all  other labels  into {\it  non-highway}. 
Our {\it KITTI-road-type} dataset consists of $3314$ images, $232$ of which are non-highway.We adapt AlexNet~\cite{krizhevsky2012imagenet} to deal with road-type classification. 

\begin{figure*}[t]
\vspace{-0.4cm}
\centering
\includegraphics[width=0.96\linewidth]{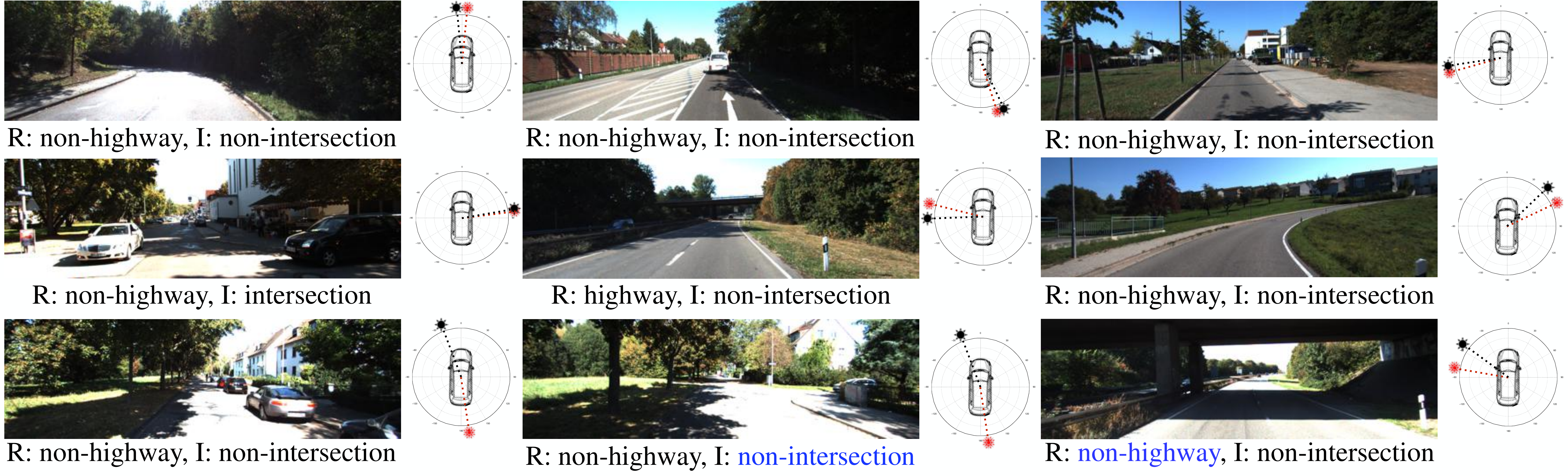}
\vspace{-0.4cm}
\caption{{\bf Qualitative results of semantics estimated by our CNNs: } predictions of road  and intersection types are under each image. The  failure prediction is marked in {\color{blue} blue}. We also show  predicted sun direction on the right of each image in black, while  ground truth  in {\color{red} red}.}
\label{fig:deep-qualitative}
\end{figure*}

%% file: method.tex
 \section{Sun and Semantics for Self-Localization }
 
 In this section we describe how to exploit OSM maps, sun direction, the presence/absence of an intersection, the road type,  the speed limit and visual odometry to perform self-localization. 

\subsection{Map Representation}
Following \cite{brubaker2013lost}, we represent the map with a directed graph, where nodes encode street segments and edges define connections between the segments. As illustrated in \figref{fig:merge-three}(b), each street segment in the map is described by its starting and ending points $\pos_0$ and $\pos_1$ respectively, initial heading angle $\streetOrient$, maximum speed limit $\streetSpeedLimit$, road type $\streetRoadType$, intersection type $\streetInterType$, and a curvature parameter $\streetCurvature$. The curvature is zero for linear street segments (\ie, $\streetCurvature = 0$) and $\streetCurvature = \frac{\streetArcAng_{1} - \streetArcAng_{0}}{\streetLen}$ for circular arc segments where $\streetArcAng_{0}$ and $\streetArcAng_{1}$ are the starting and ending angle of the arc and $\streetLen$ is the arc length of the segment. We further define the position and orientation of a vehicle in the map in terms of the street segment $\streetNode$ that the vehicle is on, the distance traveled on that street segment $\streetDist$ as well as the offset of the local street heading $\streetAng$. Through this parametric representation, we can easily obtain the global heading of the vehicle $\bar{\streetAng}$, which is given by $\bar{\streetAng} = \streetAng + \streetOrient + \streetCurvature\streetDist$.

\subsection{State-Space Model}
\label{sec:state-space-model}
The state of the model at time $t$ is defined as $\state_t = (\streetNode_t,\streetState_t)$, where $u_t$ is the street where the vehicle is driving on and  $\textbf{s}_t = (d_t, \hat{d}_{t-1}, \theta_t, \hat{\theta}_{t-1})$, with $d_t$ and $\theta_t$  the distance and local heading angle of the vehicle w.r.t. the origin of the street $u_t$. Further, $\hat{d}_{t-1}, \hat{\theta}_{t-1}$ are the distance and angle at the previous time step $t-1$ relative to current street $u_t$. 

In this work, we employ five types of observations:  sun direction, intersection type, road type, vehicle's velocity, and visual odometry. Let $\textbf{y}_t = ( \phi_t, i_t, r_t, v_t, \textbf{o}_t)$ be the observations at time $t$, with  $\phi_t$  the estimated relative sun direction with respect to vehicle's current heading, $i_t$ the estimated intersection type in front of the vehicle, $r_t$ the  estimated type of road the vehicle is currently driving on,  $v_t$ the measured velocity of the vehicle (calculated from the visual odometry), and $\textbf{o}_t$ the visual odometry measurement. We assume that the observations are conditionally independent given the state $\state_t$, and define a fully factorized observation model:
\begin{align}
p(\textbf{y}_t | \mvec{x}_t) = p(\phi_t | \mvec{x}_t)p(i_t | \mvec{x}_t)p(r_t | \mvec{x}_t)p(v_t | \mvec{x}_t)p(\mvec{o}_t | \mvec{x}_t)
\label{eq:likelihood}
\end{align}
We now describe each likelihood term in more details follow by the state transition distribution.

\vspace{0.2cm}
\noindent{\bf Sun Direction:} This term encourages the estimate of the sun direction computed from the car's location and time of the day to agree with the estimated sun direction with our Sun-CNN. 
Towards this goal, we model the sun direction as a Gaussian distribution:
\begin{align}
p(\phi_t | \textbf{x}_t) = {\mathcal N}(\phi_t | \mu_\phi(\state_t), \Sigma^{s}),
\label{eq:sun-likelihood}
\end{align}
where $\mu_\phi(\state_t)$ is a deterministic function which computes the relative sun position from the vehicle's current state $\state_t$ and the global sun position that calculated from time of the day and coarse geo-location. Please refer to appendix for a detailed formulation of $\mu_\phi(\state_t)$. We learn the covariance $\Sigma^{s}$  from the training set.

\vspace{0.2cm}
\noindent{\bf Intersection Type:} This term captures the fact that the presence/absence of an intersection in a road segment should agree with the estimation of  our Intersection-CNN. As mentioned in \secref{sec:intersection-classification}, we define an intersection to be visible if it lies  between (6.25m-23m) and it intersects with the viewing frustum. Following this definition, we further partition the OSM street segments that contain an intersection  into three sub-segments as shown in \figref{fig:merge-three}(c). The first one is connected to the intersection and has length  6.25m. As it is too close to the intersection, we assume that the intersection cannot be seen. The second segment spans from (6.25m-23m) and the intersection is visible. The third segment is more than 23m away from the intersection and thus is labeled as not containing an intersection. Thanks to  this partitioning,  the intersection type does not vary within a street segment. We thus  model this observation as a Bernouille distribution only affecting the street segment ID: 
\begin{align}
p(i_t | \mvec{x}_t) = p(i_t | u_t) = \gamma^{\delta^{int}_{i_t,u_t}} (1-\gamma)^{1-\delta^{int}_{i_t,u_t}},
\label{eq:inter-likelihood}
\end{align}
with $\delta^{int}_{i_t,u_t}$ a delta function with value 1 if the presence of an intersection in $u_t$ is the same as the one output by the intersection-CNN, and 0 otherwise. We estimated $\gamma$ using the confusing matrix of the training data. In practice we use $\gamma=0.8$.

\vspace{0.2cm}
\noindent{\bf Road Type:} This term captures the fact that roads of the same type as the one estimated by our Road-Type-CNN should have higher likelihood. 
We thus model it with a Bernouille distribution:
\begin{align}
p(r_t | \mvec{x}_t) = p(r_t | u_t) =  \beta^{\delta^{type}_{r_t,u_t}} (1-\beta)^{1-\delta^{type}_{r_t,u_t}},
\label{eq:road-likelihood}
\end{align}
with $\delta^{type}_{r_t,u_t}$ a delta function with value 1 if the type of $u_t$ is the same as the one output by the Road-CNN, and 0 otherwise.
We estimated $\beta=0.9$ using the confusing matrix of the training data.

\vspace{0.2cm}
\noindent{\bf Speed Limit:} This term models the fact that the speed of driving depends on the speed-limit. 
We model the probability of the vehicle's velocity as a piece-wise constant function, which depends on the road type. If the vehicle's velocity is less than the maximum speed, we treat it as a uniform distribution over all possible speeds. We employ as maximum speed   25km/h over the speed-limit, as we empirically  observed in KITTI that the driving speed is very often above the speed limit. A uniform distribution makes sense as there might be traffic ahead of us, slowing our driving. 
If the velocity exceeds the maximum speed, we give very low probability. Thus
\begin{align}
p(v_t | \mvec{x}_t) = p(v_t | u_t) = 
\begin{cases} 
\frac{0.99}{\streetSpeedLimit_{u_t} + V_0} , & \text{if}\ v_t \leq \streetSpeedLimit_{u_t} + V_0 \\
\epsilon , & \text{otherwise}
\end{cases}
\label{eq:velocity-likelihood}
,
\end{align}
where $\streetSpeedLimit_{u_t}$ denotes the maximum speed  of street $u_t$ and $\epsilon$ represents a very small number, which we set to  $10^{-4}$. 

\vspace{0.2cm}
\noindent{\bf Odometry:} Following  \cite{brubaker2013lost} we model  odometry  as a Gaussian distribution linear in $\textbf{s}_t$, 
\begin{align}
p(\textbf{o}_t | \textbf{x}_t) = {\mathcal N}(\textbf{o}_t | \mmat{M}_{u_t}\textbf{s}_t, \Sigma^{\mvec{o}}_{u_t}),
\label{eq:odom-likelihood}
\end{align}
with $\mmat{M}_{u_t} = [\mvec{m}_d, \mvec{m}_\theta]^T$, $\mvec{m}_d = (1, -1, 0, 0)^T$ and $\mvec{m}_\theta = (\alpha_u, -\alpha_u, 1, -1)^T$. Since the visual odometry performs significantly worse at higher speeds, we learn different variances for highways and city/rural roads as suggested by \cite{brubaker2013lost}. $\Sigma^{\mvec{o}}_{u_t}$ is therefore a function of the street $u_t$ and is directly learned from data as discussed in \secref{sec:parameter-learning}.

\begin{table}[t]
\vspace{-0.5cm}
\centering
\begin{center}
\scalebox{0.8}{
\begin{tabular}{|c||c|c|c|}
\hline
Accuracy &Non-intersection &Intersection &Total\\
\hline
Intersection-CNN & 82.8$\%$ & 75.29$\%$ & 81.62$\%$\\
\hline
Human Perception & 86.62$\%$ & 79.53$\%$ &$85.51\%$\\
\hline
\hline
Accuracy &Non-highway &Highway &Total\\
\hline
Road Type-CNN &99.45$\%$ & 91.38 $\%$ & 98.88$\%$\\
\hline
Human Perception &99.51$\%$ & 93.5 $\%$ & 99.06$\%$\\
\hline 
\end{tabular}}
\end{center}
\vspace{-0.3cm}
\caption{Accuracy for Intersection-CNN and Road-Type-CNN vs humans.}
\vspace{-4mm}
\label{tab:road-inter-accuracy}
\end{table}

\vspace{0.2cm}
\noindent{\bf State Transition:}
We factorize the state transition distribution  
as $$p(\state_t | \state_{t-1}) = p(\streetNode_t | \state_{t-1}) p(\streetState_t | \streetNode_t, \state_{t-1}).$$

As is common in the literature, we use a second order constant velocity model to describe the vehicles' motion represented by linear transition dynamics corrupted with Gaussian noise for $p(\streetState_t | \streetNode_t, \state_{t-1})$.
We define the probability of changing streets, 
$p(\streetNode_t | \state_{t-1})$, as a sigmoid, encoding the fact that transitions between street segments are more likely to occur as we arrive to the end of the street segment. We refer the 
 readers to  the appendix for more details. \

\subsection{Inference}
Inference in our model consists of recursively computing the distribution $p(\mvec{x}_t | \mvec{y}_{1:t})$.
The posterior can be factored as $p(\state_{t} | \obs_{1:t}) = p(\streetState_{t} | \streetNode_{t}, \obs_{1:t}) p(\streetNode_{t} | \obs_{1:t})$ using the product rule, where $p(\streetNode_{t} | \obs_{1:t})$ is a discrete distribution over streets and $p(\streetState_{t} | \streetNode_{t}, \obs_{1:t})$ is a continuous distribution over the position and orientation on a given street.
The discrete distribution $p(\streetNode_{t} | \obs_{1:t})$ is easily represented as a multinomial distribution over street labels.
As for the continuous distribution $p(\streetState_{t} | \streetNode_{t}, \obs_{1:t})$, we represent it with a Gaussian mixture model. With our model assumptions, the filtering distribution can then be written recursively as
\begin{equation}
p(\state_t | \obs_{1:t}) = \frac{p(\obs_{t}|\state_t) \int p(\state_{t}|\state_{t-1}) p(\state_{t-1}|\obs_{1:t-1}) d\state_{t-1}}{p(\obs_{t}|\obs_{1:t-1})}.
\end{equation}
Here, we use the efficient approximation of  \cite{brubaker2013lost} to approximate our posterior.
This is possible as by design our likelihood terms are either Gaussian or only dependent on the street segment which is a discrete variable. More details on inference can be found in the appendix.

\subsection{Learning}
\label{sec:parameter-learning}

\begin{table*}[t]
\vspace{-0.4cm}
\centering
\scalebox{0.9}{
\begin{tabular}{|c||c|c|c|c|}
\hline
&Localization Time &Computational Time &Gini Index &Success Rate\\
\hline
\hline
Brubaker \etal \cite{brubaker2013lost} &$46 \pm 24$s &$0.69$s/frame &$0.765 \pm 0.057 $ &$81.8\%$\\
Our Full Model &${\bf 25 \pm 21}$s &${\bf 0.48}$s/frame &${\bf 0.879 \pm 0.013}$ &${\bf 90.9\%}$\\
\hline
\end{tabular}
}
\vspace{-0.3cm}
\caption{{\bf Comparison against \cite{brubaker2013lost}}. Our approach significantly outperforms \cite{brubaker2013lost} in term os localization time, computation time, Gini index and success rate. }
\label{tab:compare-marcus}
\end{table*}

In this section, we discuss how we learn the odometry and sun direction noise, $\Sigma_{\streetNode}^{\mvec{o}}$  and $\Sigma^{s}$.  As both odometry and sun direction are modeled as a Gaussian distribution, we apply maximum likelihood estimation (MLE) to learn the variance from the data. For odometry, we compute the ground truth by projecting each image onto the road using  GPS+IMU, and calculating the odometry of the GPS. For sun direction, we employ the images in the \emph{KITTI-Sun} dataset.We note that different parameters were learned for highways and city/rural roads as the visual odometry performs significantly worse at higher speeds. Leave-one-out cross validation was used to learn the parameters.

%% file: results.tex
\section{Experimental Evaluation}
\label{sec:experiment}

We validate the effectiveness of our approach on the training sequences from the KITTI visual odometry benchmark \cite{Geiger2012CVPR} where ground truth trajectory is available.

\begin{figure*}[t]
\vspace{-0.5cm}
\centering
\includegraphics[width=0.98\linewidth]{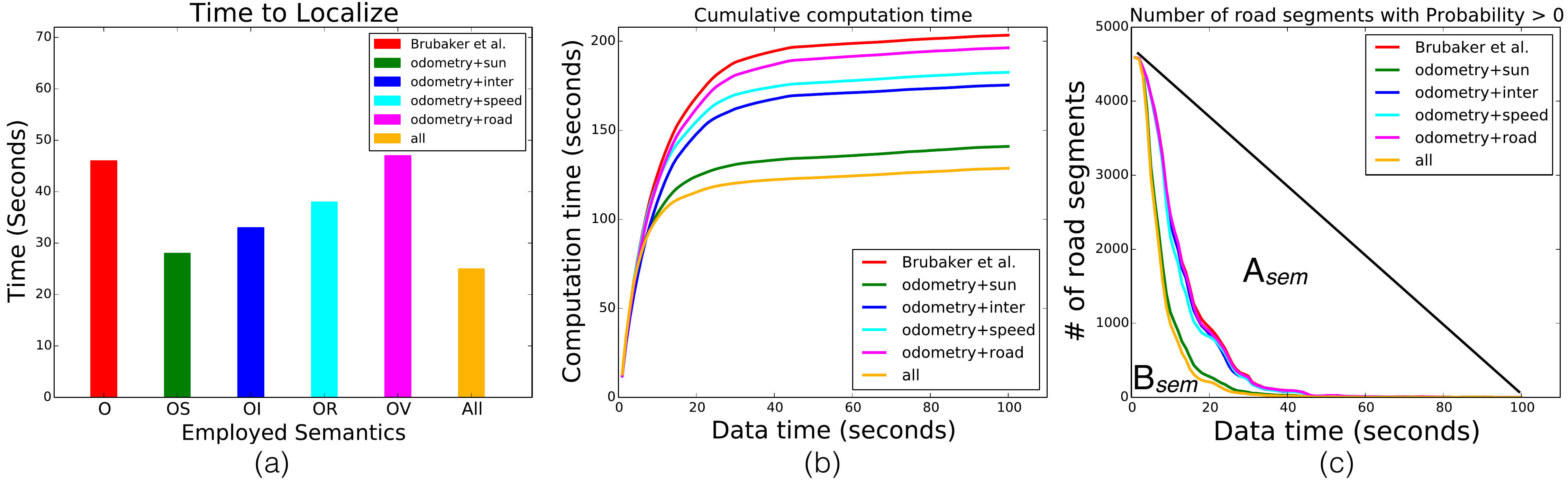}
\vspace{-0.4cm}
\caption{{\bf Quantitative evaluation of our localization approach: }(a) {\bf Average localization time.} The average localization time when employing different semantic cues. It is computed across the sequences that all methods can localize (b) {\bf Cumulative computation time.} The average computation time when employing different semantic cues. We run our program on 8 cores with a basic Python implementation. Although our inference is slower than real-time at first, we argue that we can reduce it by employing more computational resource (\textit{i.e.} using more cores). (c) {\bf Road segments with probability larger than 0.} The average number of road segments (3m) with probability larger than 0 when employing different semantic cues. We denote the area below each curve as $B_{sem}$, while the area between the black line and the curve as $A_{sem}$, where $sem$ denotes the semantic cues employed. In our scenario, the closer $\frac{A_{sem}}{A_{sem}+B_{sem}}$  to $1$, the better. We compute the average using only the sequences that all methods localize.}
\label{fig:time-gini}
\vspace{-4mm}
\end{figure*}

\subsection{Deep Learning for semantics}
\label{sec:exp-deep}
For all three semantic tasks, we hold out images from each odometry sequence to create each split, and use the rest for training, resulting in 11 models. We augment the training images 10 times by cropping/mirroring. All hyper-parameters are selected via grid search on the validation set. Note that sequence 03 is excluded from the semantics evaluation as GPS and IMU data is not available.

\vspace{-0.2cm}
\paragraph{Sun-CNN:}
We initialized our network with weights pre-trained on ImageNet~\cite{deng2009imagenet}, and employ a learning rate of  $10^{-5}$ for fc7,  and $10^{-6}$ for the rest. Training a deep net for this task took around 3 hours requiring approximately 200 epochs to converge.  We use the cumulative sun direction error (angle between prediction and ground truth direction) across the \emph{KITTI-Sun} dataset as metric. We compare our model against the state-of-the-art sun estimation approach of Lalonde \etal \cite{lalonde2009estimating}. We use default values for all parameters in \cite{lalonde2009estimating}, except the sun visibility $v_s$ and the horizontal line $v_h$. The sun visibility probability $P(v_s)$ of all images are set to 0.8, since KITTI was collected in good weather. The horizontal line $v_h$ is computed using KITTI calibration. For more details, we refer the readers to \cite{lalonde2009estimating}. To verify our current settings best describe the geodesics of the rotation manifold, we also exploit different loss functions and parameterization. We first change the distance metric between two vectors from cosine distance to L2 distance, which we denoted as \textit{L2-Vec}. We further re-parameterize the sun direction in the image with a single variable (angle) and modify the network structure to a single continuous output, which we denote as \textit{L2-Angle}. \figref{fig:sun-curve} shows that our Sun-CNN significantly outperforms \cite{lalonde2009estimating} by a large margin and our choice of loss function and parameterization consistently surpass those of others. Over $60\%$ of the images have prediction errors less than $20$ degrees with our approach, while only  $25\%$ for \cite{lalonde2009estimating}.  A few qualitative results are shown in \figref{fig:deep-qualitative}. We also visualized the fc7 embedding space learned by Sun-CNN with t-SNE \cite{van2008visualizing} in \figref{fig:tSNEsun}. Sun-CNN not only separates images based on their sun direction, but also preserves the relationship among images, \ie, images with similar sun direction are embedded closer.


To further understand why our Sun-CNN performs so well, we visualize  in \figref{fig:imagenet-vs-sun-cnn} the activation map of some units at different layers of the network as well as  a network with same architecture but trained on the ImageNet classification task. Unlike classic network whose deep convolution layers learn high-level concepts \cite{zhou2015object}, our kernels at conv3 and conv4 layers start to capture the illumination variations in the images, e.g. shadows and shadings. We argue that this is because our Sun-CNN learns that shading and shadows are crucial for predicting sun direction. As a proof of concept,  \figref{fig:sun-activation} shows the activations of several neurons in conv3 and conv4 layers. Despite being trained on a \emph{high-level} task (\ie,  sun direction), our Sun-CNN automatically learns to fire on both shadings and shadows.

\vspace{-0.2cm}
\paragraph{Intersection-CNN:} For each fold, we uniformly oversample the images labeled as intersections so that each class is balanced. Since we are trying to differentiate whether an image is taken at a \textit{place} where an intersection is visible, we  pre-trained the convolutional net on the MIT Places dataset~\cite{zhou2014learning}. We set the learning rate of the last fully connected layer to $6 \times 10^{-4}$, and the rest to $6 \times 10^{-5}$. Training took 2 hours and  150 epochs to converge. As shown in  \tabref{tab:road-inter-accuracy}  our Intersection-CNN can achieve $82\%$ accuracy. To validate human performance on this task, we also asked $3$ in-house annotators to provide labels for each image.  \tabref{tab:road-inter-accuracy} shows that human perception is only $4\%$ higher than our deep model, demonstrating the difficulty of this task. Some qualitative results are shown in \figref{fig:deep-qualitative}.

\vspace{-0.2cm}
\paragraph{Road-Type-CNN:}
As the data is unbalanced, we  oversampled the less frequent classes. We pre-trained our network  on the MIT Places dataset~\cite{zhou2014learning}. The learning rate of fc7 is set to $10^{-2}$, while we use $10^{-3}$ for the rest. On average, training took around 1.5 hours and approximately 100 epochs to converge.  As shown in  \tabref{tab:road-inter-accuracy}  the Road-Type-CNN can achieve almost $99\%$ performance. We also validate human performance on this task by asking $3$ in-house annotators to label each image. The performance of the Road-Type-CNN is comparable to human perception.

\vspace{-0.2cm}
\subsection{Localization}
\label{sec:exp-localization}
We subsample the KITTI sequences from 10 frame/s to 1 frame/s. Slower rates were found to suffer from excessive accumulated odometry error, while a higher rate may suffer from correlations in semantic errors. We utilize LIBVISO2 \cite{geiger2011stereoscan} to compute visual odometry and velocity. For sun direction, presence of an intersection, and road-type, we employ the corresponding CNN models trained in the split that does not contain the test sequence. Note that sequences in KITTI have different routes, the training/test images are thus spatially non-overlapping as well. We tested our localization algorithm on both the full sized map of Karlsruhe, as well as the sequence-specific, cropped maps provided by \cite{brubaker2013lost}. The cropped maps include the region which contains the ground truth trajectory and, on average, cover an area of 2 km$^2$ and contain 100 km of drivable roads.  While \cite{brubaker2013lost} pruned dirt roads and alleyways during preprocessing stage, we preserve all drivable roads in the map, making the localization problem more difficult (see appendix for comparison).

\begin{table*}[t]
\centering
\begin{tabular}{|ccccc||c|c|c|c|c|c|c|c|c|c||c|}
\hline
O & S & I & R & V & 00  & 01  & 02  & 03 &05  & 06  & 07  & 08  & 09  & 10  & Average  \\
\hline
\hline
\checkmark &  & & &  & 22s & 90s & 27s & 36s & 52s & * & 27s & {\bf 79}s & 30s & 43s &$46 \pm 24$s \\
\checkmark & \checkmark  & & &  & {\bf 20}s & 16s & 22s & 27s & {\bf 43}s & 69s & {\bf 13}s & {\bf  79}s & 26s & 33s & $28 \pm 22$s  \\
\checkmark &  & \checkmark  & &  & 22s & 23s & 25s & 30s & 51s & 68s & 34s & 80s & 18s & 41s & $33 \pm 21$s  \\
\checkmark &  & & \checkmark  &  & 23s & 23s & 27s & 36s & 52s & * & 27s & {\bf 79}s & 30s & 42s &$38 \pm 18$s\\
\checkmark &  & & & \checkmark  & 22s & 90s & 28s & 34s & 53s & * & 27s & {\bf 79}s & 36s & 42s & $47 \pm 24$s \\
\checkmark & \checkmark  & \checkmark & \checkmark & \checkmark  & {\bf 20}s & {\bf 12}s & {\bf 21}s & {\bf 21}s & {\bf 43}s & {\bf 31}s & {\bf 13}s & 80s & {\bf 13}s & {\bf 15}s & ${\bf 25 \pm 21}$s  \\
\hline
\hline
\end{tabular}
\caption{{\bf Quantitative evaluation:}  "O", "S", "I", "R", "V" represent the observation types that were used during  inference, i.e., visual odometry, sun direction, intersection type, road type and velocity. The average localization time, position and heading error are computed with sequences that localizes. Sequences that did not localize are indicated with a "*". No approach localize Sequence 04. Full table with all combinations of semantics and all metrics is shown in appendix. When employing all semantics our approach localizes faster.}
\label{tab:quantitative}
\end{table*}

As shown in \tabref{tab:compare-marcus} our model using all semantic cues  performs much better than \cite{brubaker2013lost} in terms of  localization time, computation time and localization success rate  \footnote{We re-emphasize that the localization accuracy of \cite{brubaker2013lost} is already at level of precision of OSM and cannot be improved further. The accuracy of our approach is similar to \cite{brubaker2013lost}. Please refer to appendix for detailed numbers.}. A sequence is defined to be {\em localized} when, for at least ten seconds, there is a single mode in the posterior. Once the criteria for localization is met, all subsequent frames are considered localized. We note that we localize in sequences where \cite{brubaker2013lost} failed \footnote{One should note that as we employed a more complex map comparing to \cite{brubaker2013lost}, the results of using merely visual odometry may differ from their paper.}. 
To further evaluate how  \emph{fast} our approach can reduce the solution space, we also employ \textit{Gini index} \cite{gastwirth1972estimation}, a well-known measure of statistical dispersion, as a new metric. \figref{fig:time-gini}(b) shows the  average number of road segments (3m) that have probability larger than 0 across all sequences when employing different semantic cues.  We denote the area below each curve as $B_{sem}$, while the area between the black line and the curve as $A_{sem}$, where $sem$ denotes the semantic cues employed. The Gini index is defined as $\frac{A_{sem}}{A_{sem}+B_{sem}}$. The closer the ratio is to 1, the faster the solution space is pruned (the better).

Examples of results on  cropped maps are shown  in \figref{fig:comparison} (see appendix for results using the full map). 
We  observe that sequence 06 traverses a geometrically symmetric path, causing inference to be unable to select between the two modes using only visual odometry. When using  the sun, this ambiguity can be resolved. Since the intersections along the path are distinct, intersection information can also resolve the symmetry. 
We also show the average localization time, the average computation time and how fast an approach can reduce the solution space across all sequences in \figref{fig:time-gini}. In general, additional semantics reduce the localization time and lower the computational cost by reducing uncertainty.

\tabref{tab:quantitative} shows more detailed ablation studies.  
The symbols ``O", ``S", ``I",``R", ``V" represent the observation types that were used during inference, i.e., visual odometry, sun direction, intersection type, road type and velocity. Sequences that are not localized are indicated with a "*". In general, adding each semantic cue will help boost the performance. 
 However, if the semantic observations estimated from the images are noisy, they  might lengthen the  time required to localize. For instance, intersection-CNN hurts the localization in sequence 07 due to errors in estimating the sun direction,  while velocity slightly increase localization time due  to the over-speed of the vehicle. When combining all cues, our inference algorithm significantly improves the efficiency and achieves the best performance. One may suspect that the sun is not useful when cloudy. Note that our semantic cues are complementary and even without estimating the sun direction we still outperform \cite{brubaker2013lost} (see appendix). This is important in autonomous driving as one has to be robust.

\begin{figure*}[t]
\vspace{-0.5cm}
\centering
\setlength{\tabcolsep}{1pt} 
\begin{tabular}{ccccc}
\raisebox{60px}{\multirow{2}{*}{\includegraphics[width=0.30\linewidth]{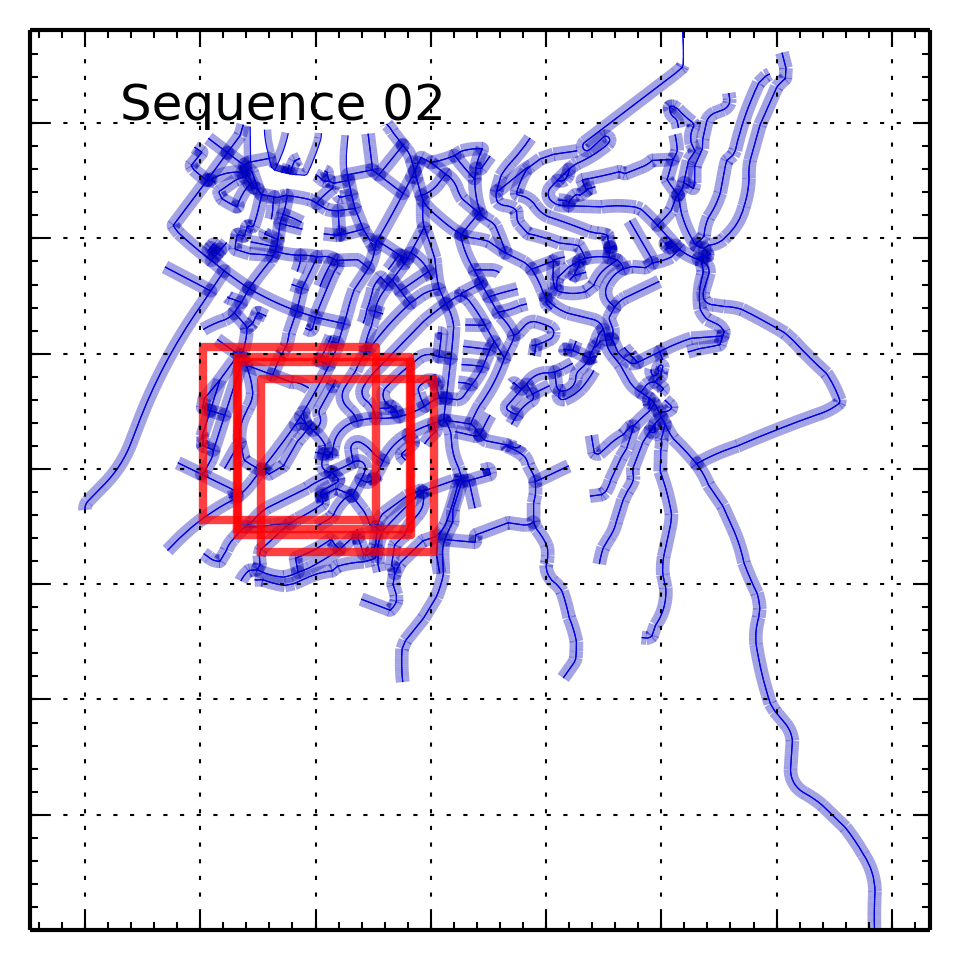}}}
&\includegraphics[width=0.16\linewidth]{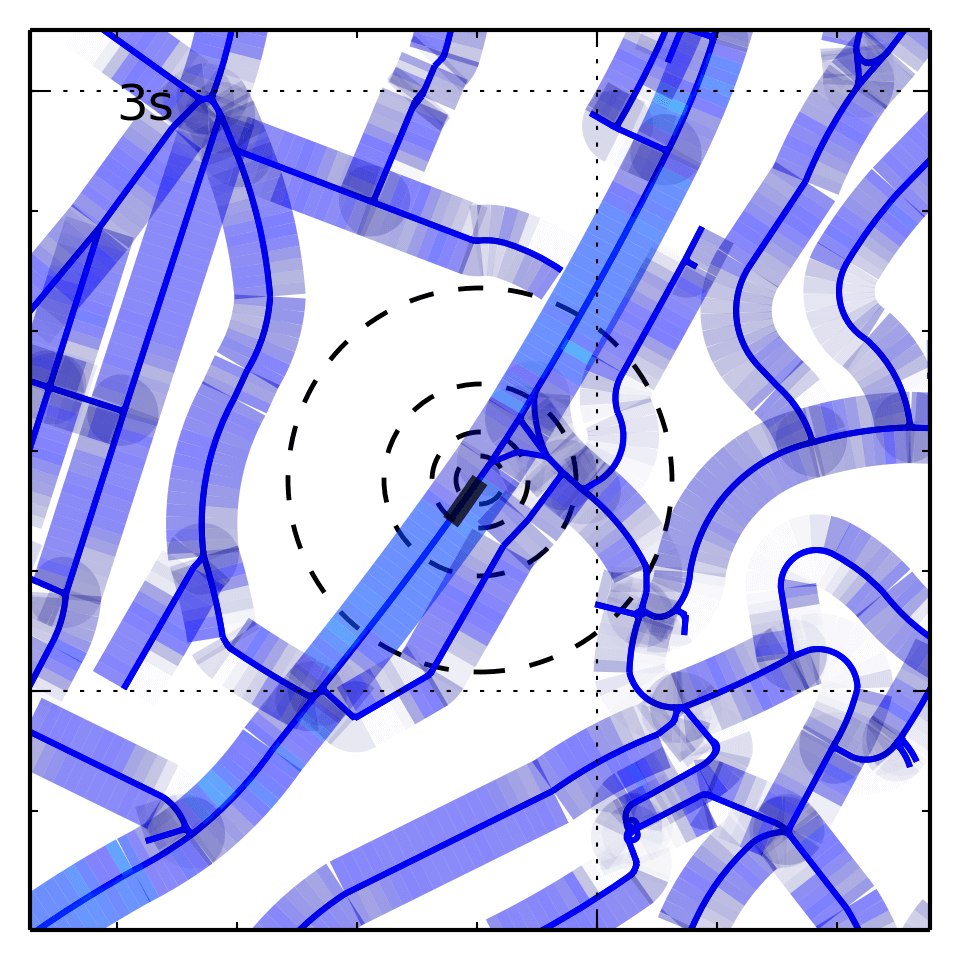}
&\includegraphics[width=0.16\linewidth]{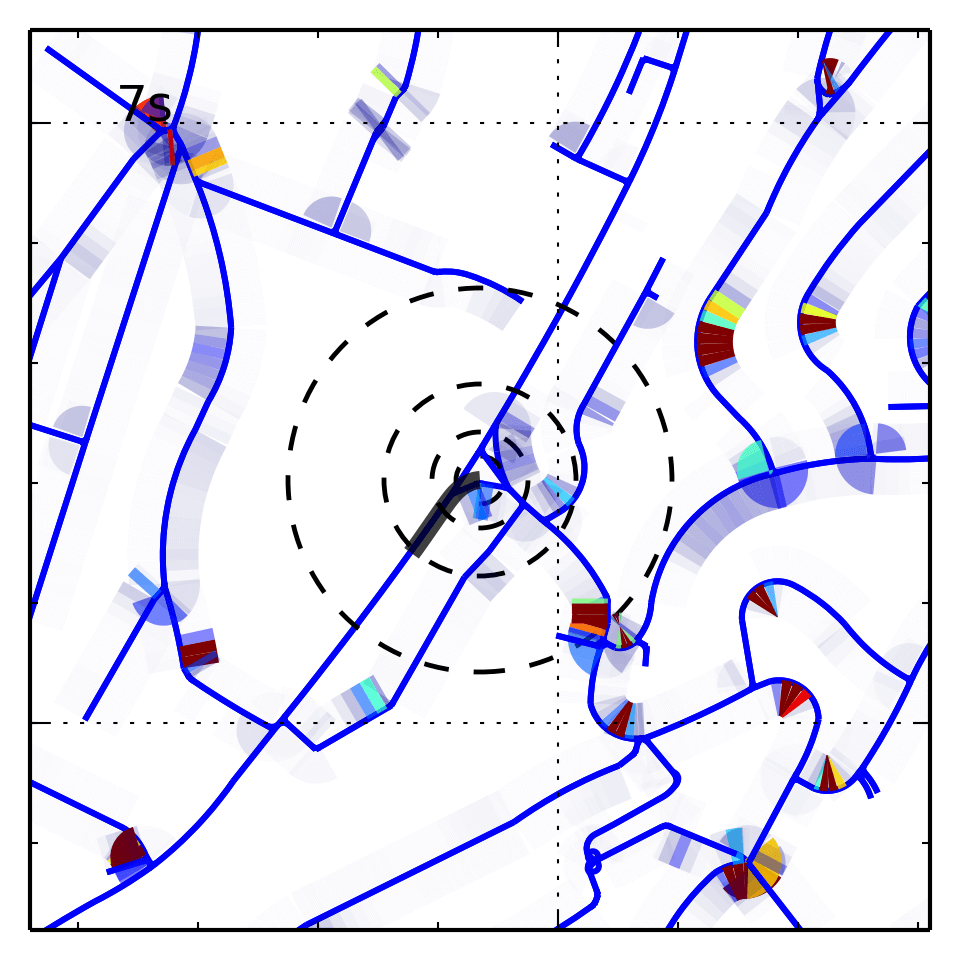}
&\includegraphics[width=0.16\linewidth]{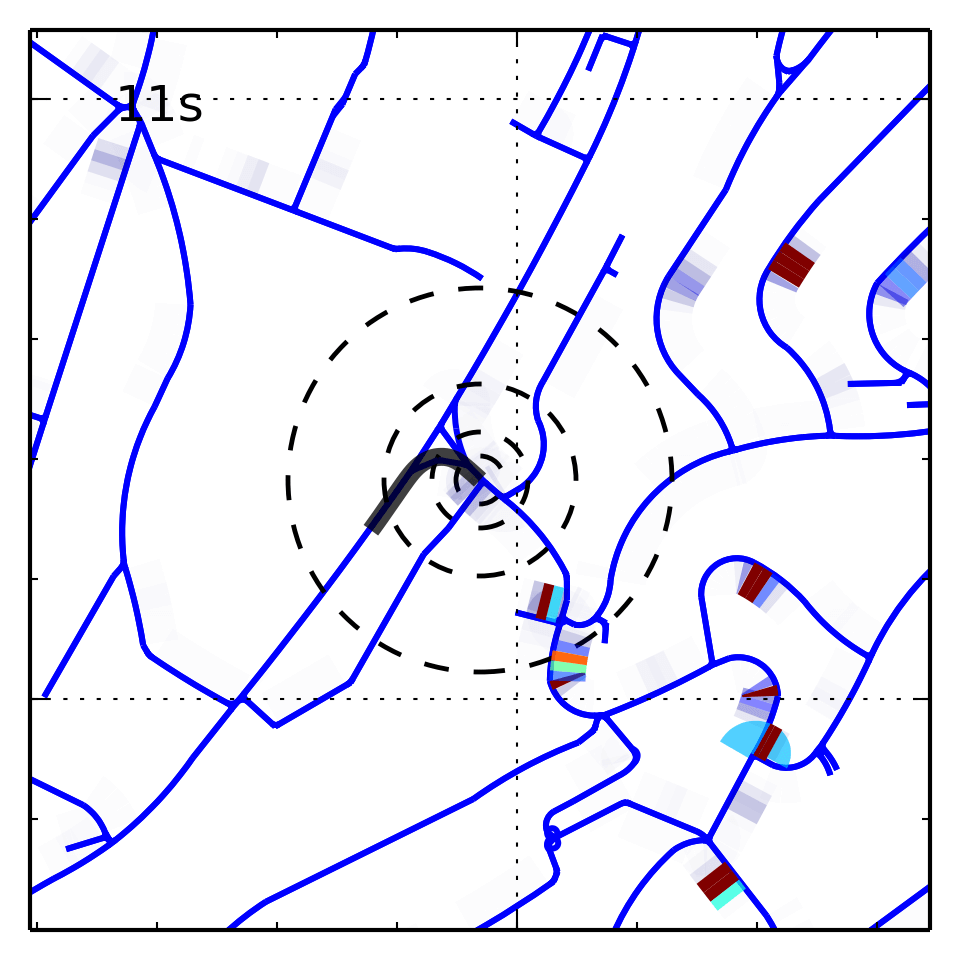}
&\includegraphics[width=0.16\linewidth]{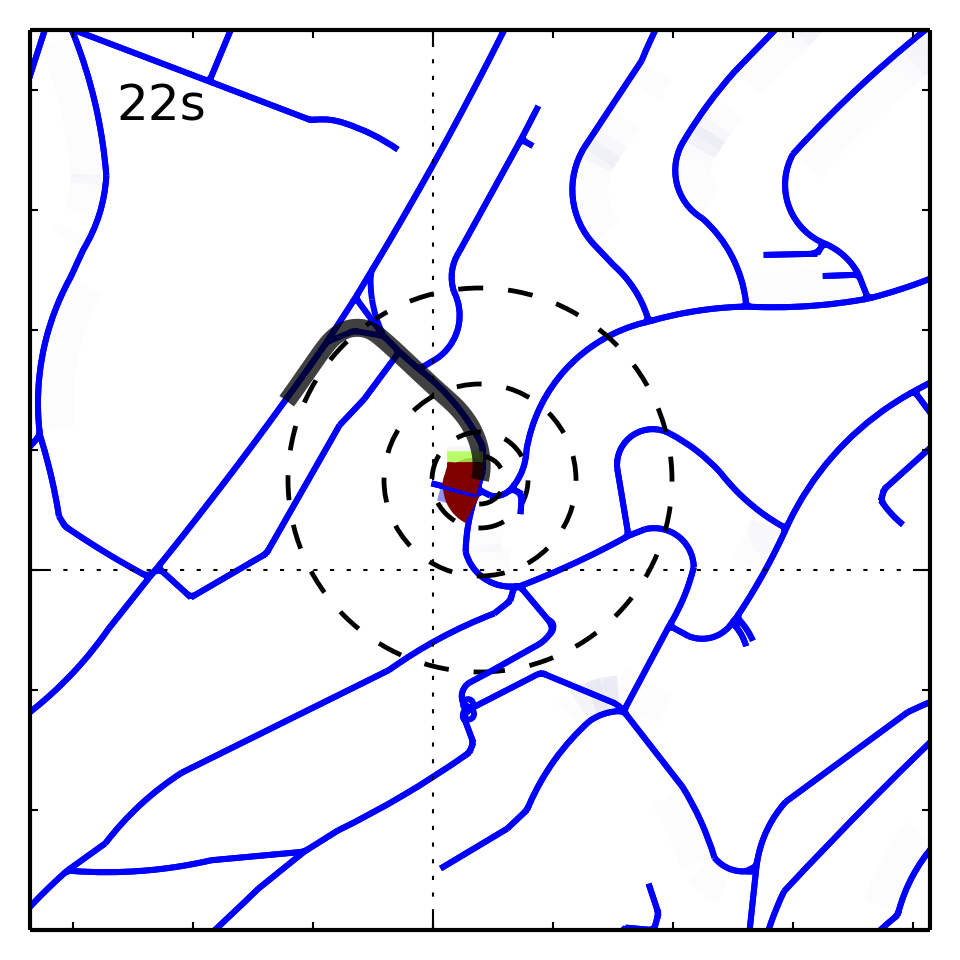}\\
&\includegraphics[width=0.16\linewidth]{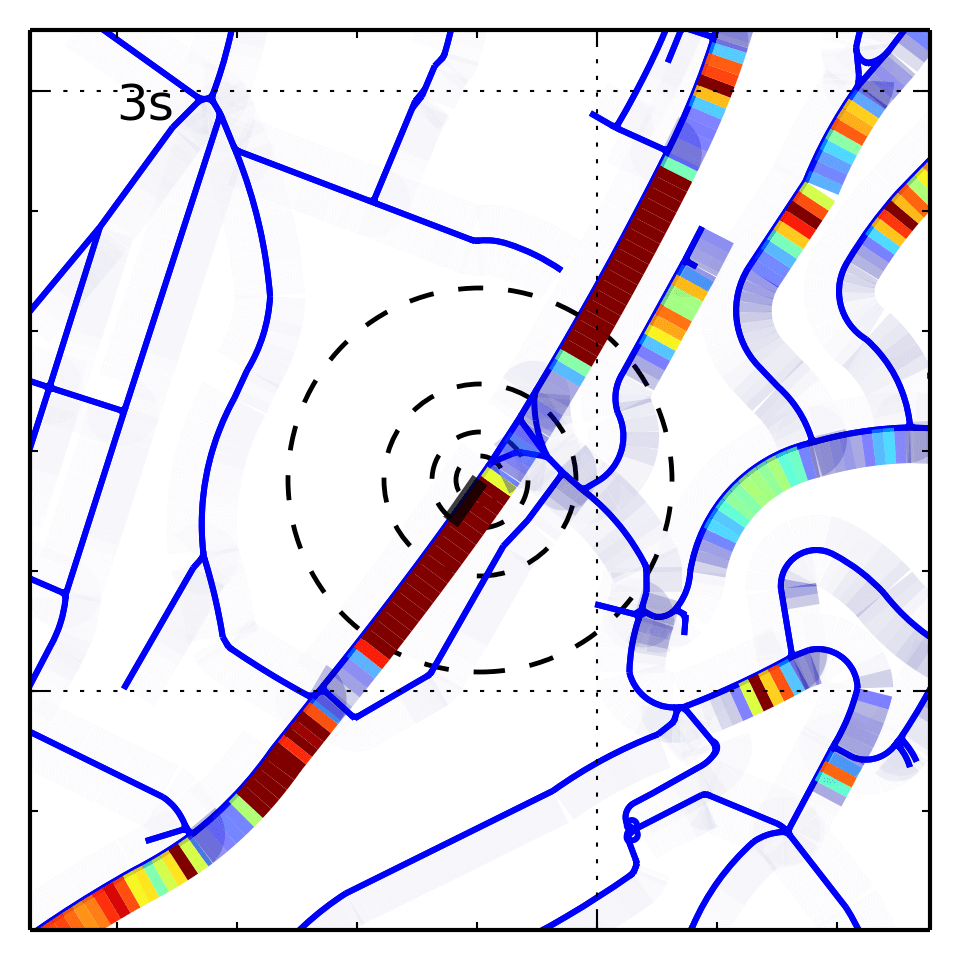}
&\includegraphics[width=0.16\linewidth]{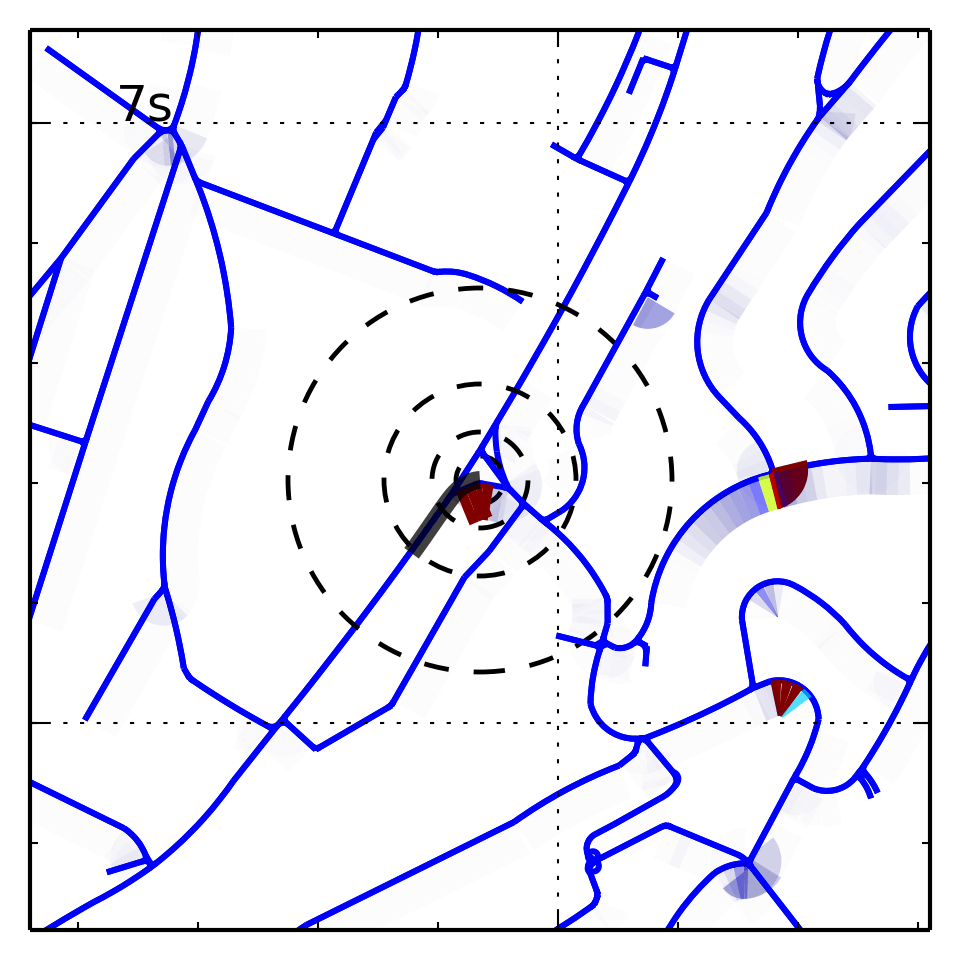}
&\includegraphics[width=0.16\linewidth]{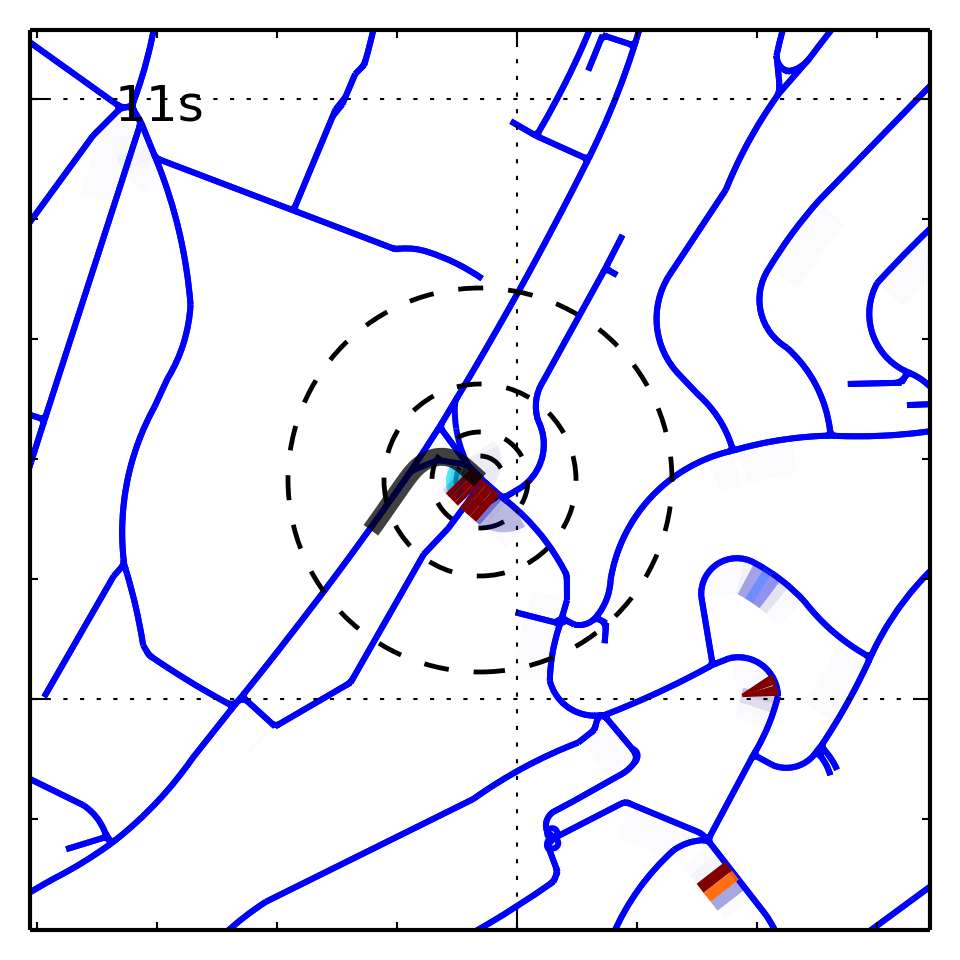}
&\includegraphics[width=0.16\linewidth]{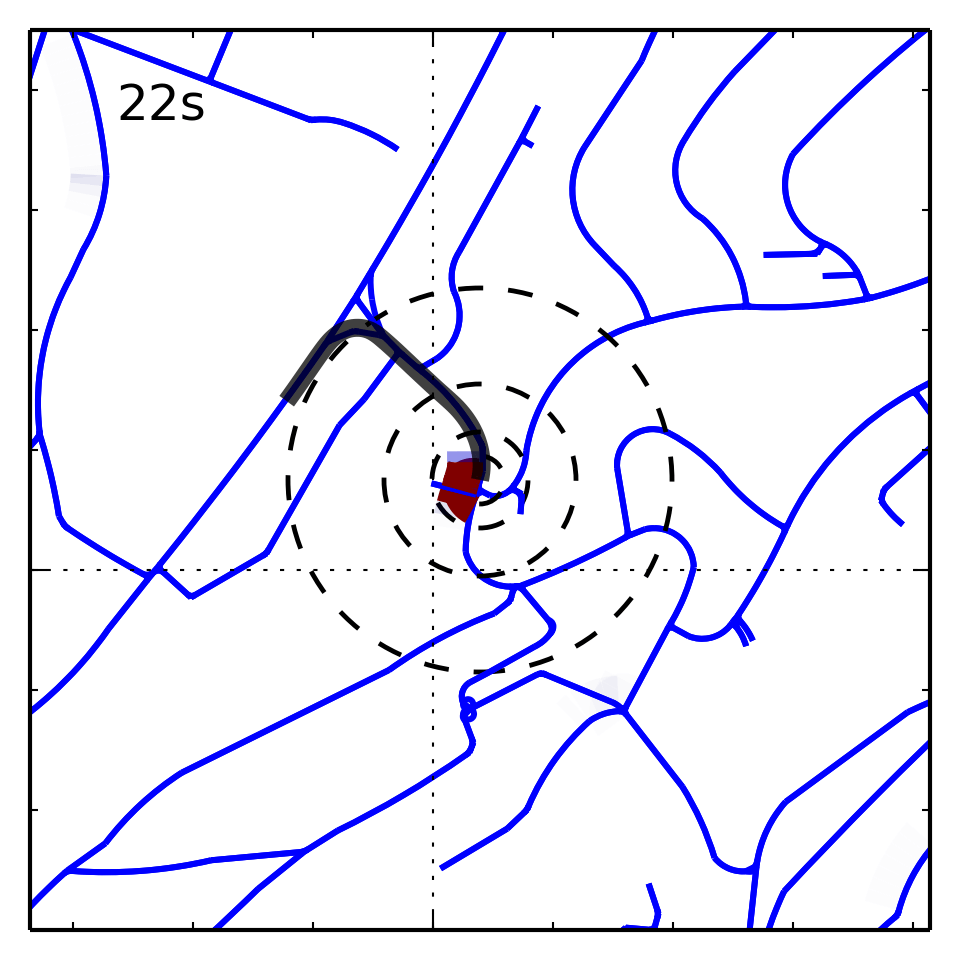}\\
\raisebox{60px}{\multirow{2}{*}{\includegraphics[width=0.3\linewidth]{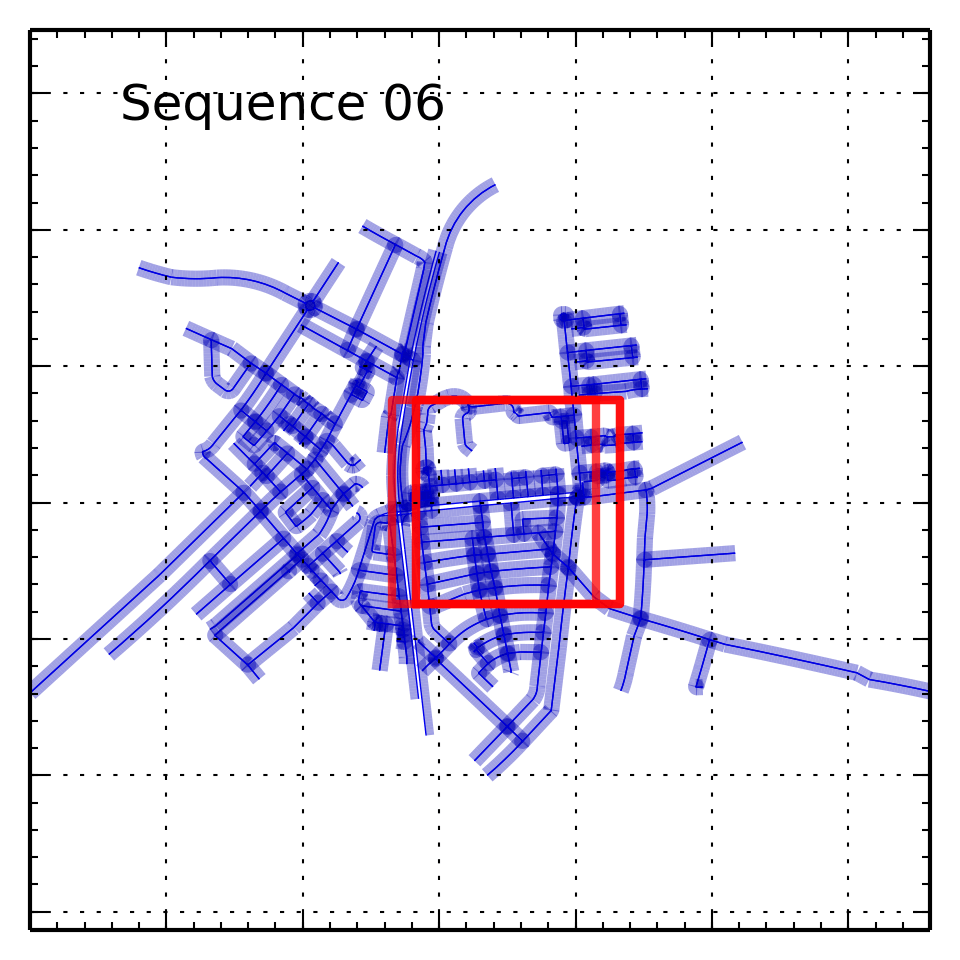}}}
&\includegraphics[width=0.16\linewidth]{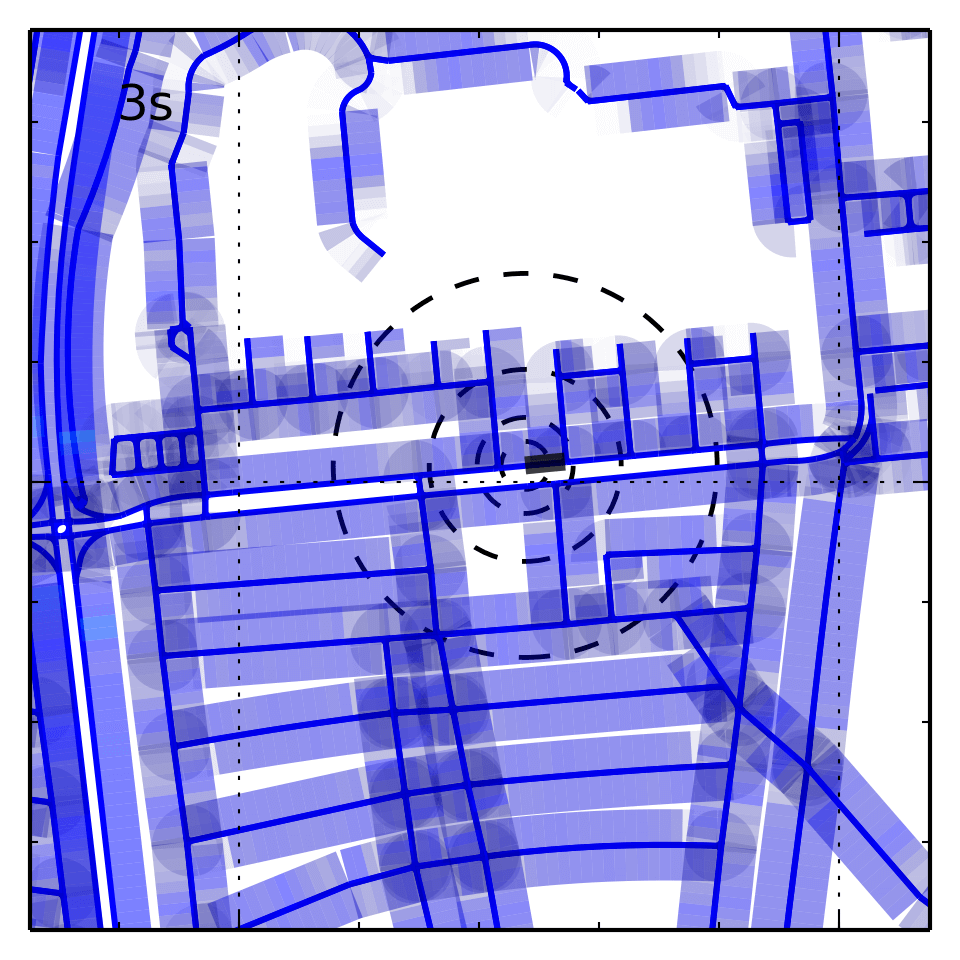}
&\includegraphics[width=0.16\linewidth]{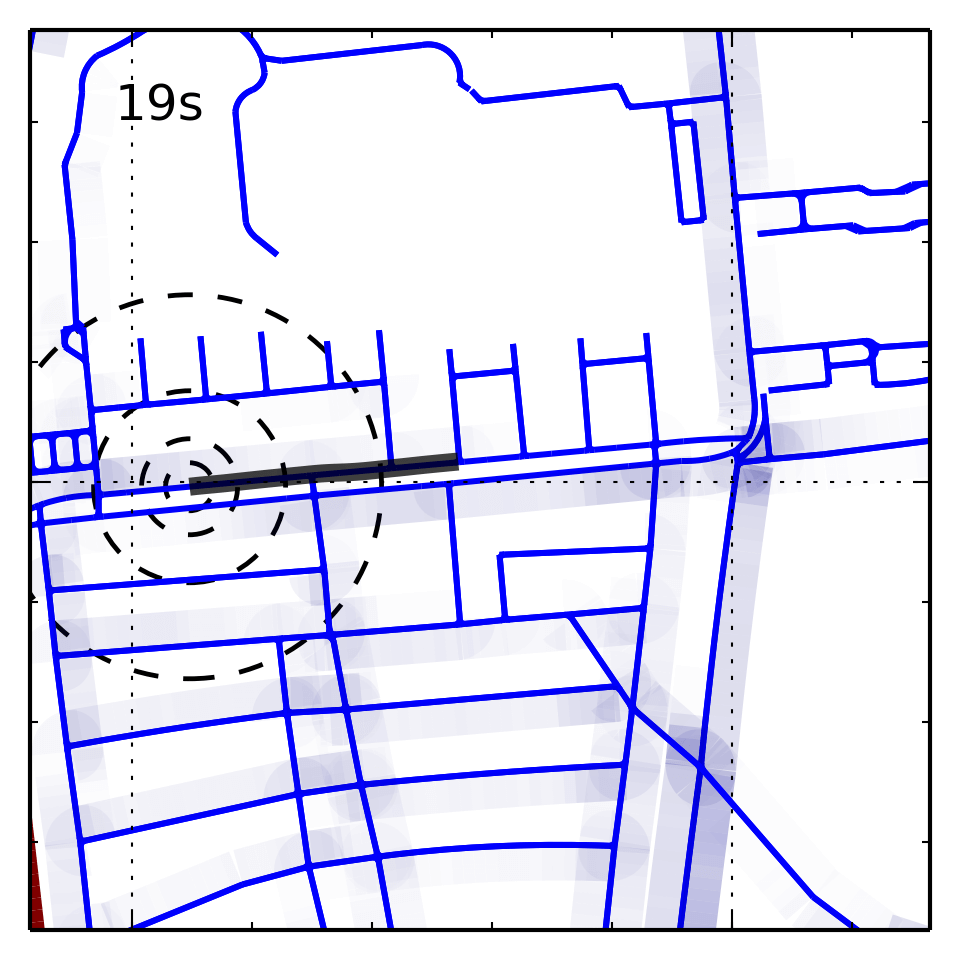}
&\includegraphics[width=0.16\linewidth]{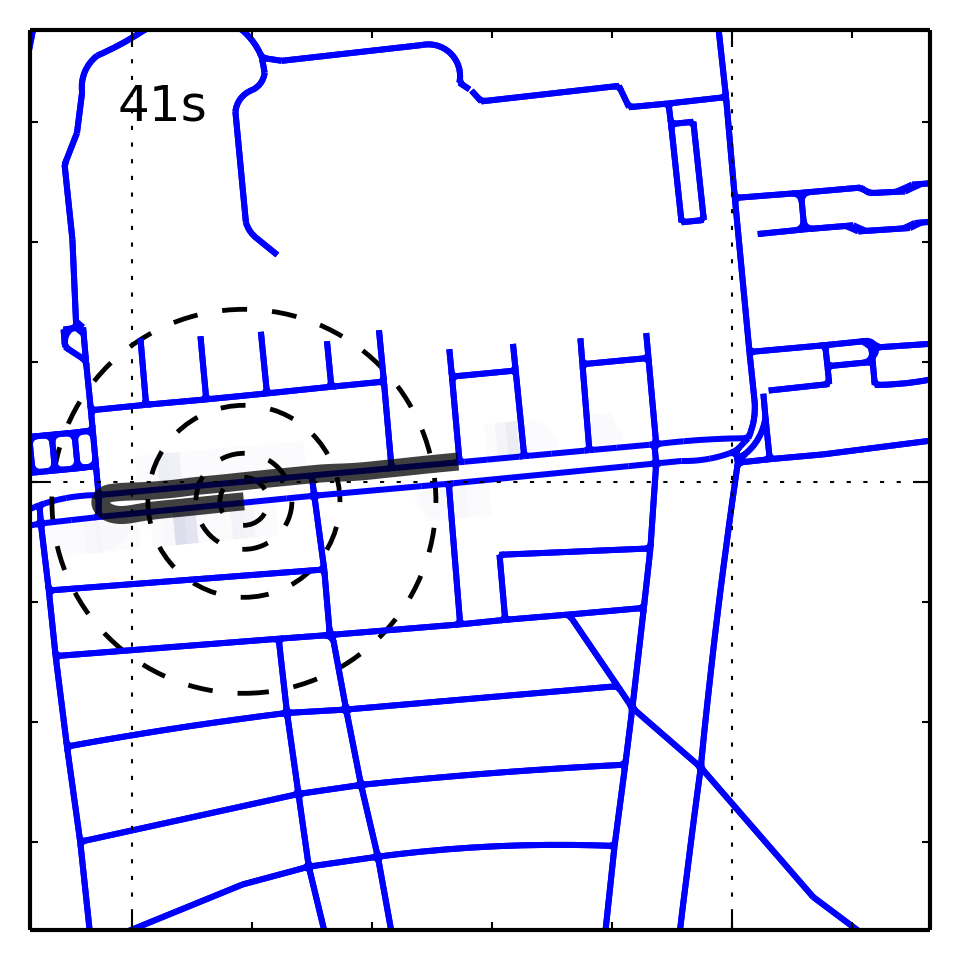}
&\includegraphics[width=0.16\linewidth]{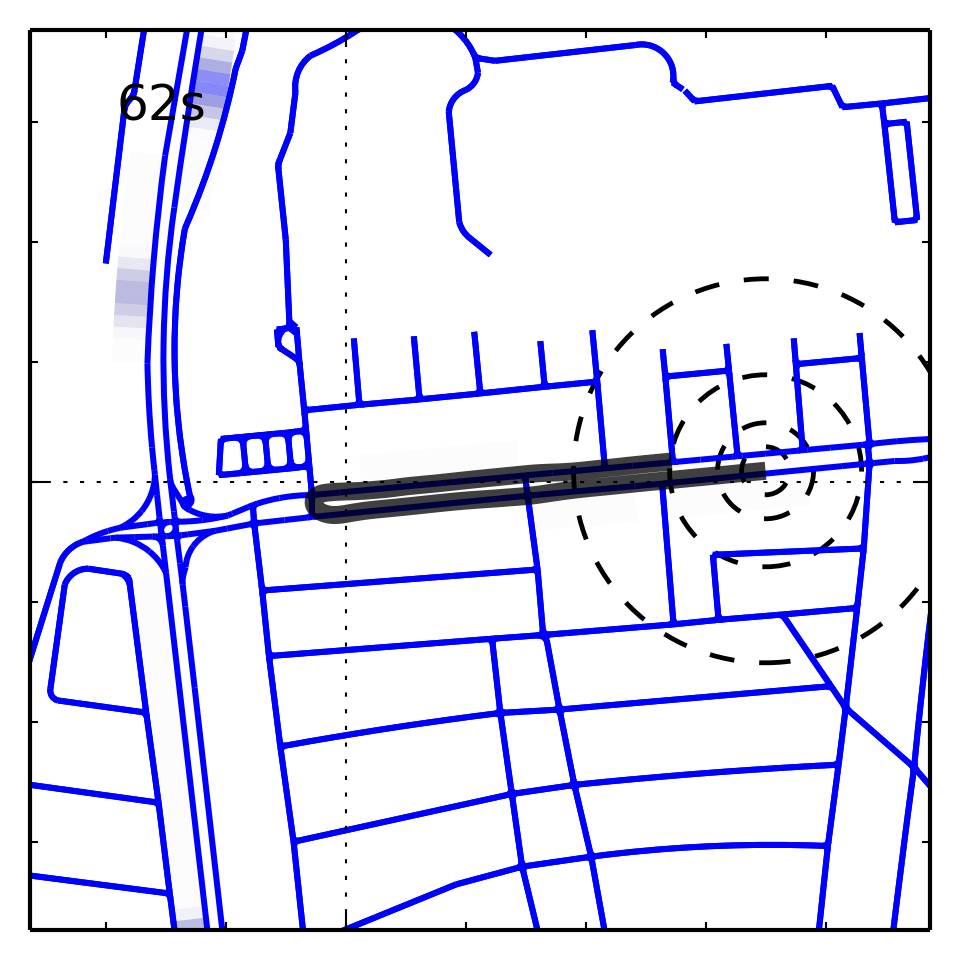}\\
&\includegraphics[width=0.16\linewidth]{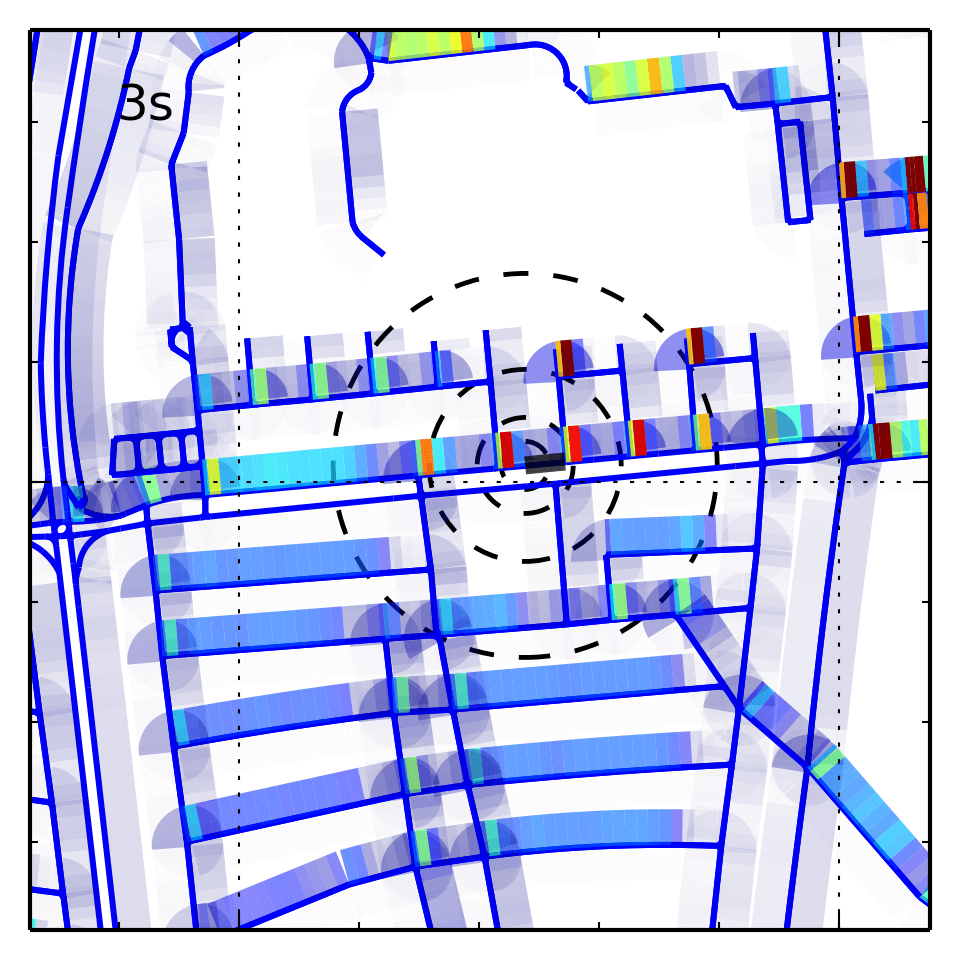}
&\includegraphics[width=0.16\linewidth]{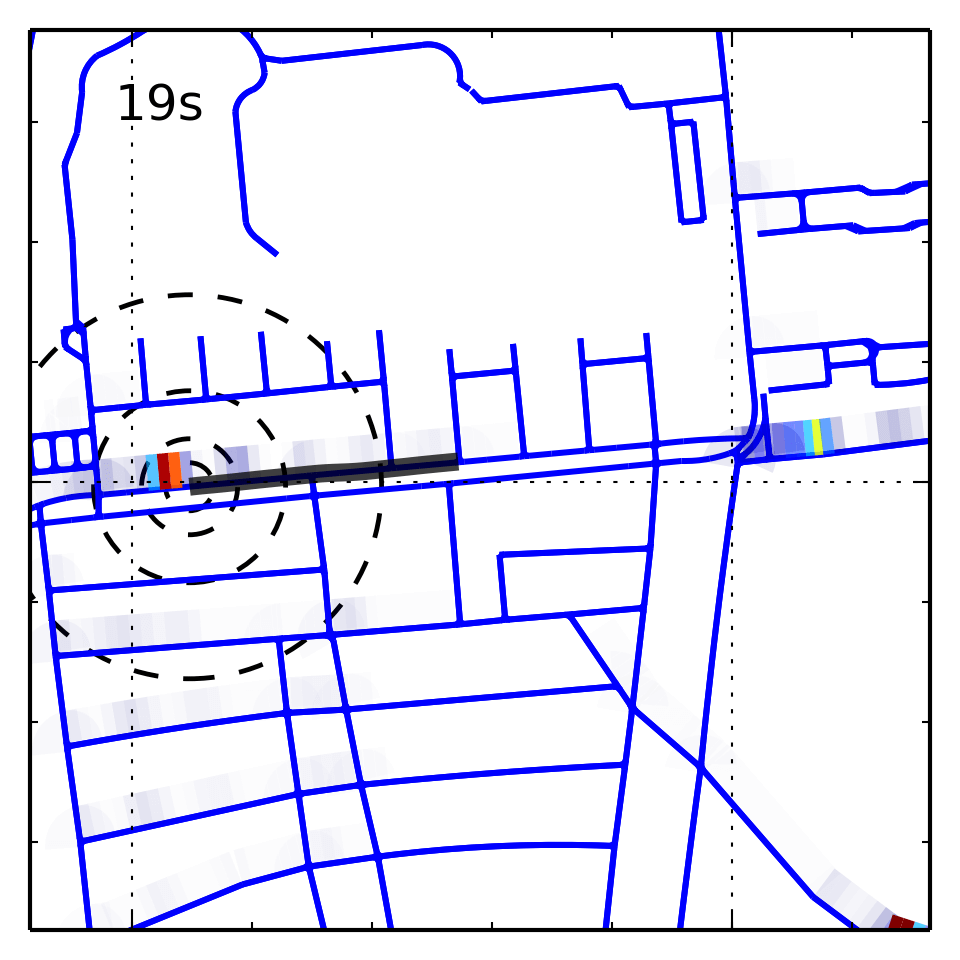}
&\includegraphics[width=0.16\linewidth]{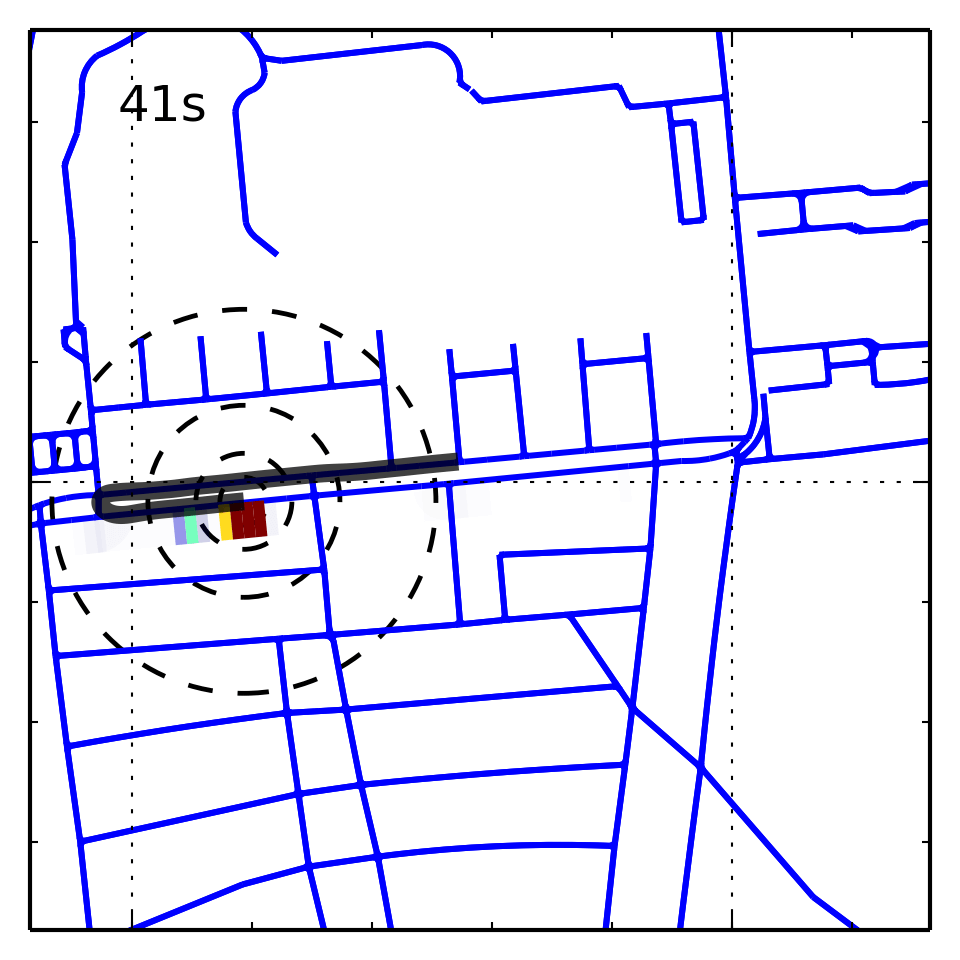}
&\includegraphics[width=0.16\linewidth]{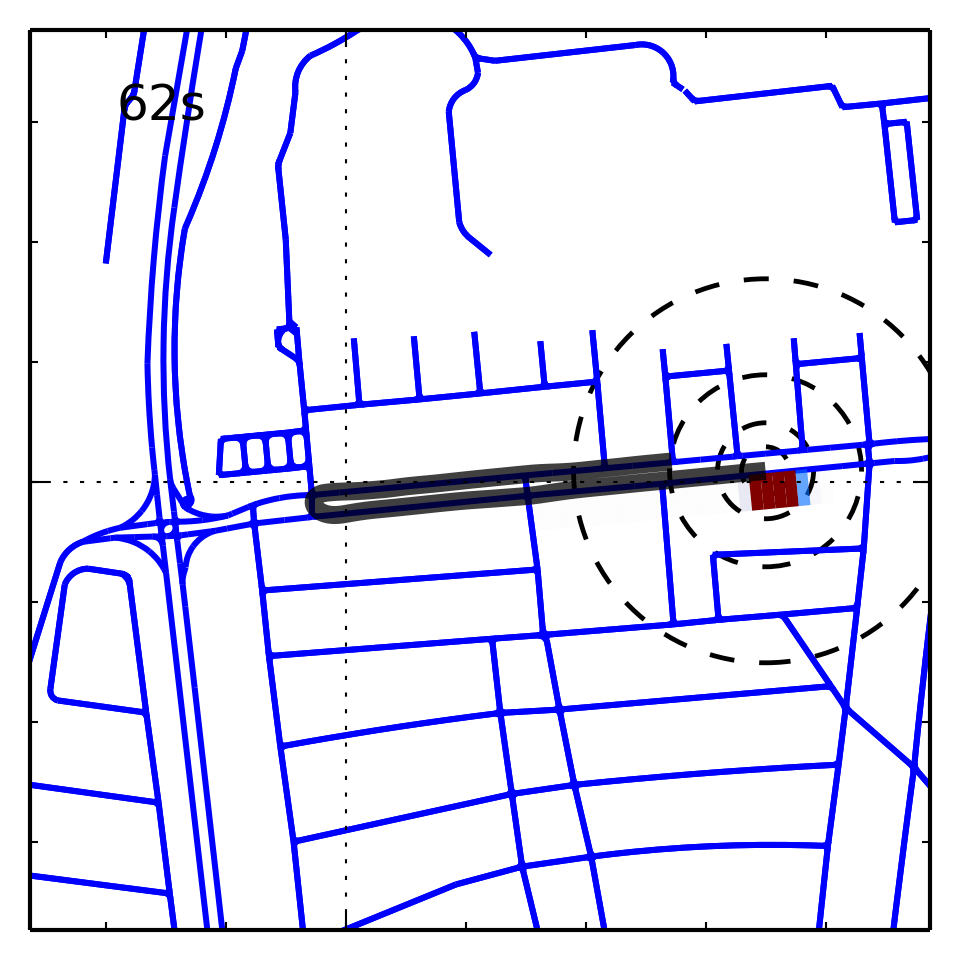}\\
\end{tabular}
\vspace{-0.3cm}
\caption{{\bf Qualitative localization results compared to \cite{brubaker2013lost}:} Results of other sequences can be found in appendix. The left most column shows the full map region for each sequence, followed by zoomed in sections of the map showing the posterior distribution over time. The upper row is the result of \cite{brubaker2013lost}, and the lower row is ours. The black line is the GPS trajectory and the concentric circles indicate the current GPS position. Grid lines are every 500m. High probability is indicated with {\color{red}red} while low probability regions are shown in {\color{blue}blue}.}
\vspace{-3mm}
\label{fig:comparison}
\end{figure*}

\begin{figure*}[tb]
\vspace{-0.5cm}
\centering
\setlength{\tabcolsep}{1pt} 
\begin{tabular}{ccccccc}
& odometry & +sun & +speed & +intersection & +road & +All\\
\raisebox{26px}{\rotatebox{90}{$t = 8$s}}
&\includegraphics[width=0.16\linewidth, trim={0 0.6cm 0 0}]{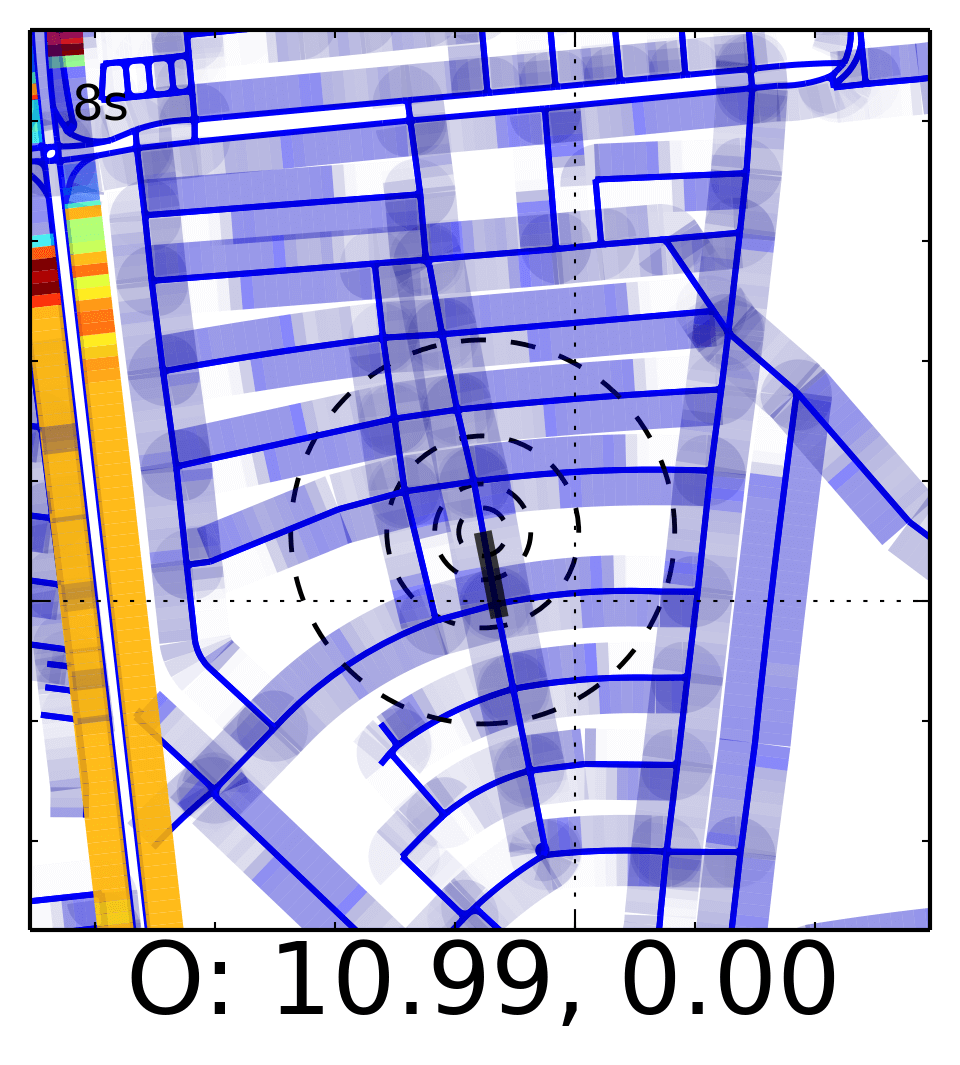}
&\includegraphics[width=0.16\linewidth,  trim={0 0.6cm 0 0}]{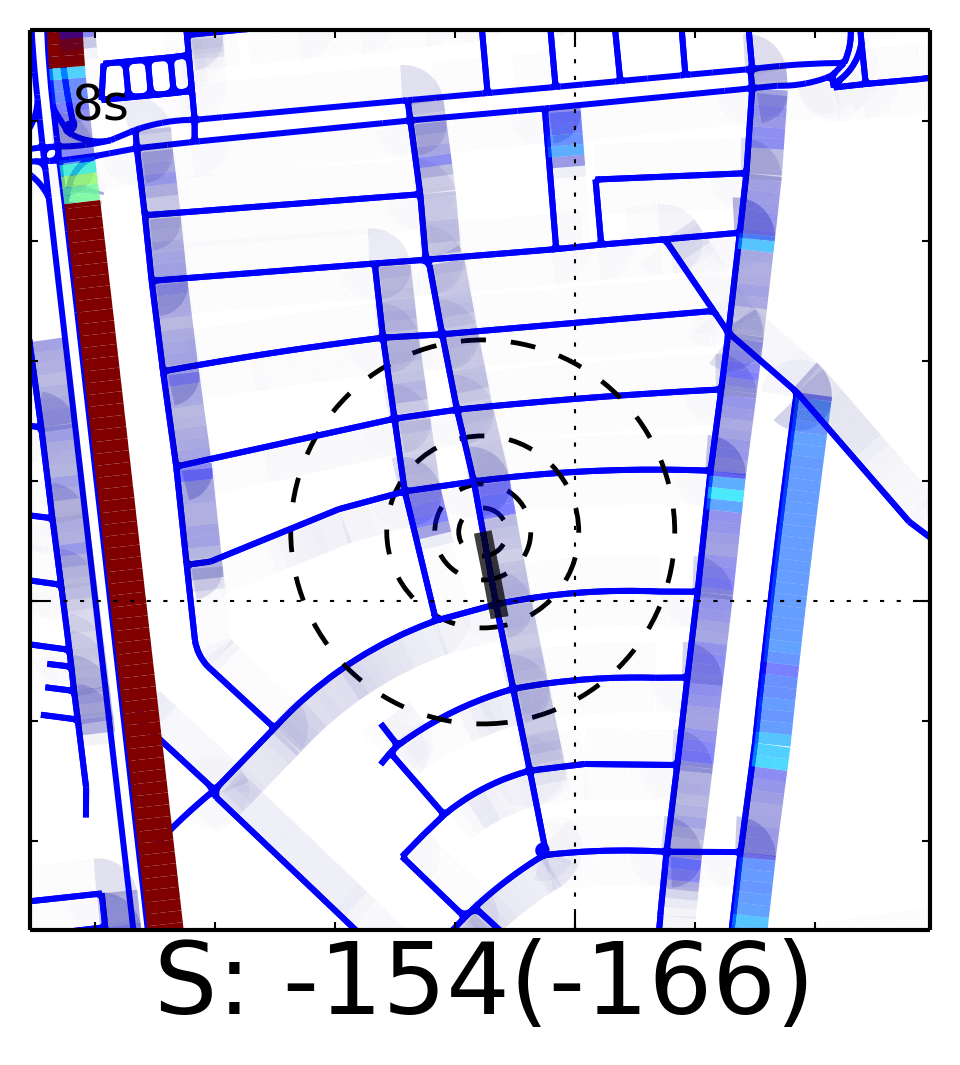}
&\includegraphics[width=0.16\linewidth,  trim={0 0.6cm 0 0}]{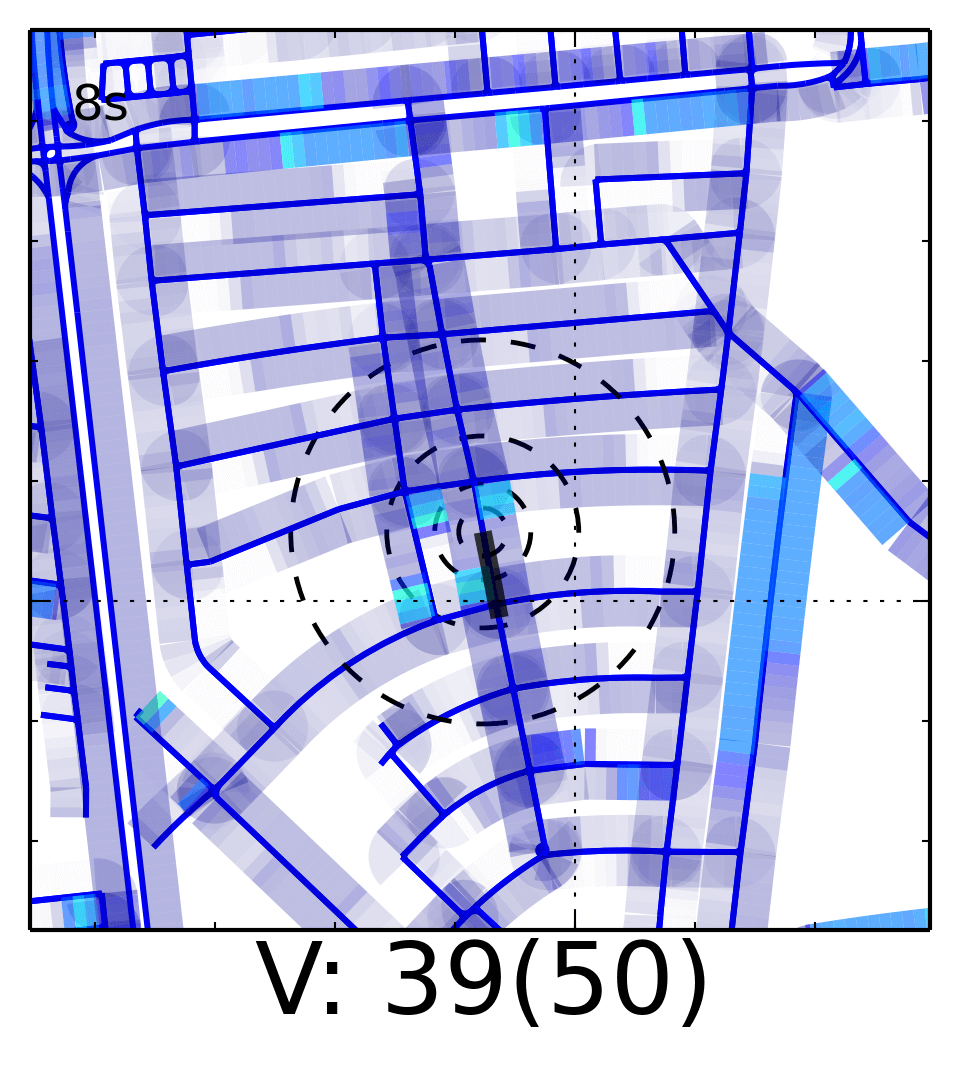}
&\includegraphics[width=0.16\linewidth,  trim={0 0.6cm 0 0}]{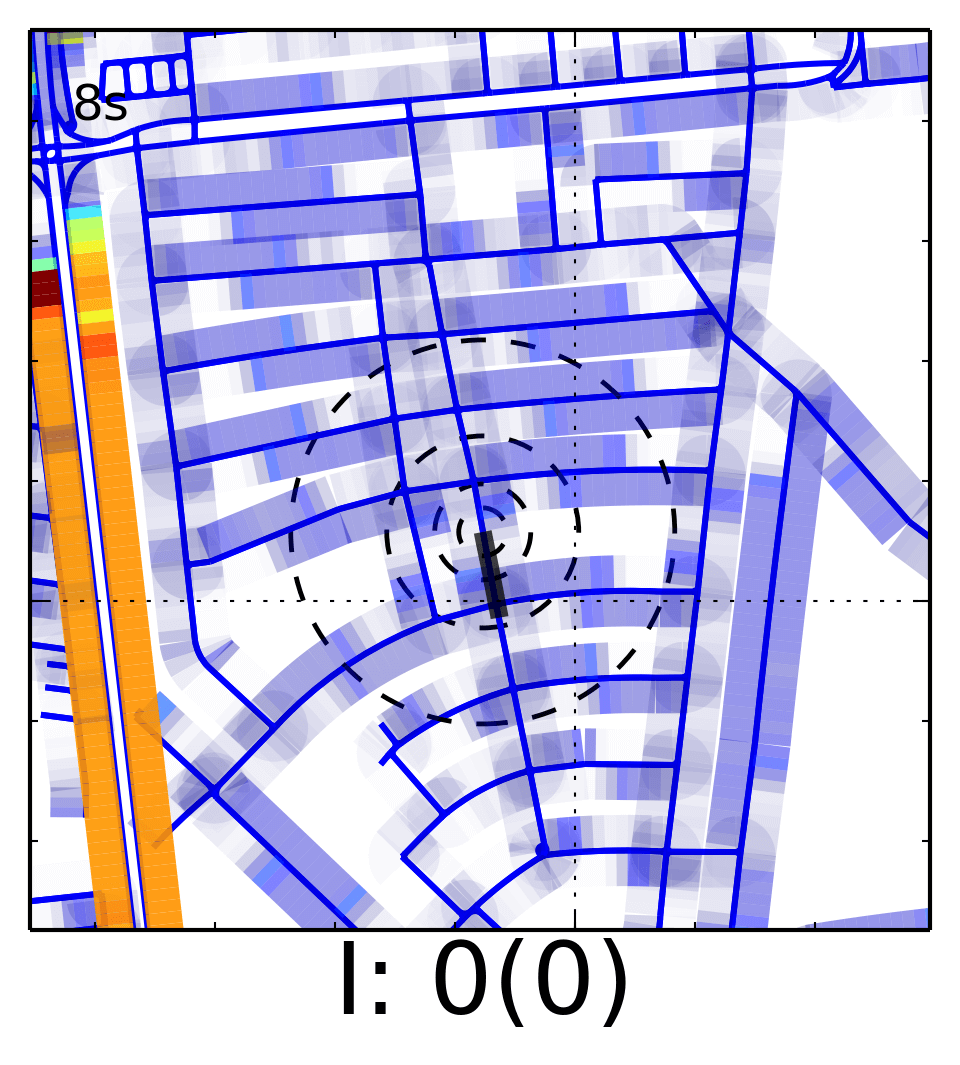}
&\includegraphics[width=0.16\linewidth,  trim={0 0.6cm 0 0}]{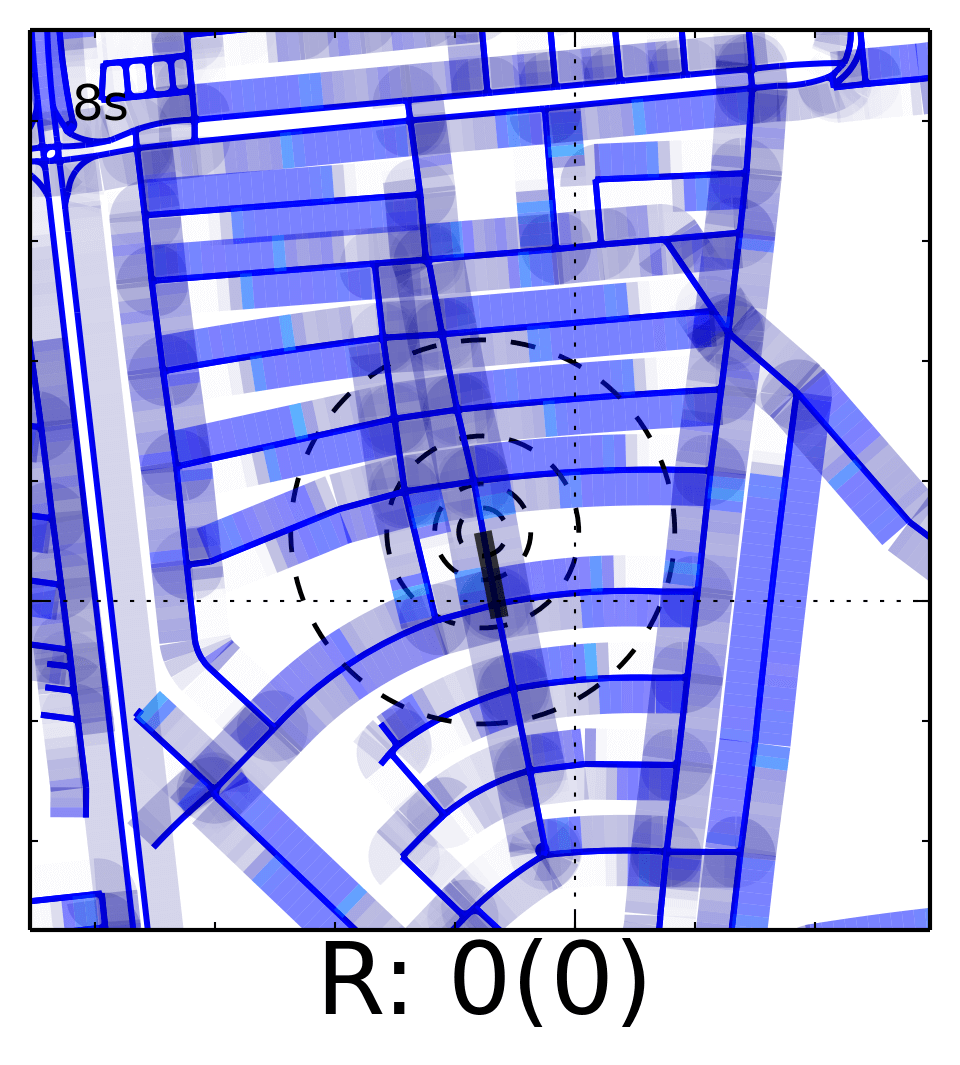}
&\includegraphics[width=0.16\linewidth,  trim={0 0.6cm 0 0}]{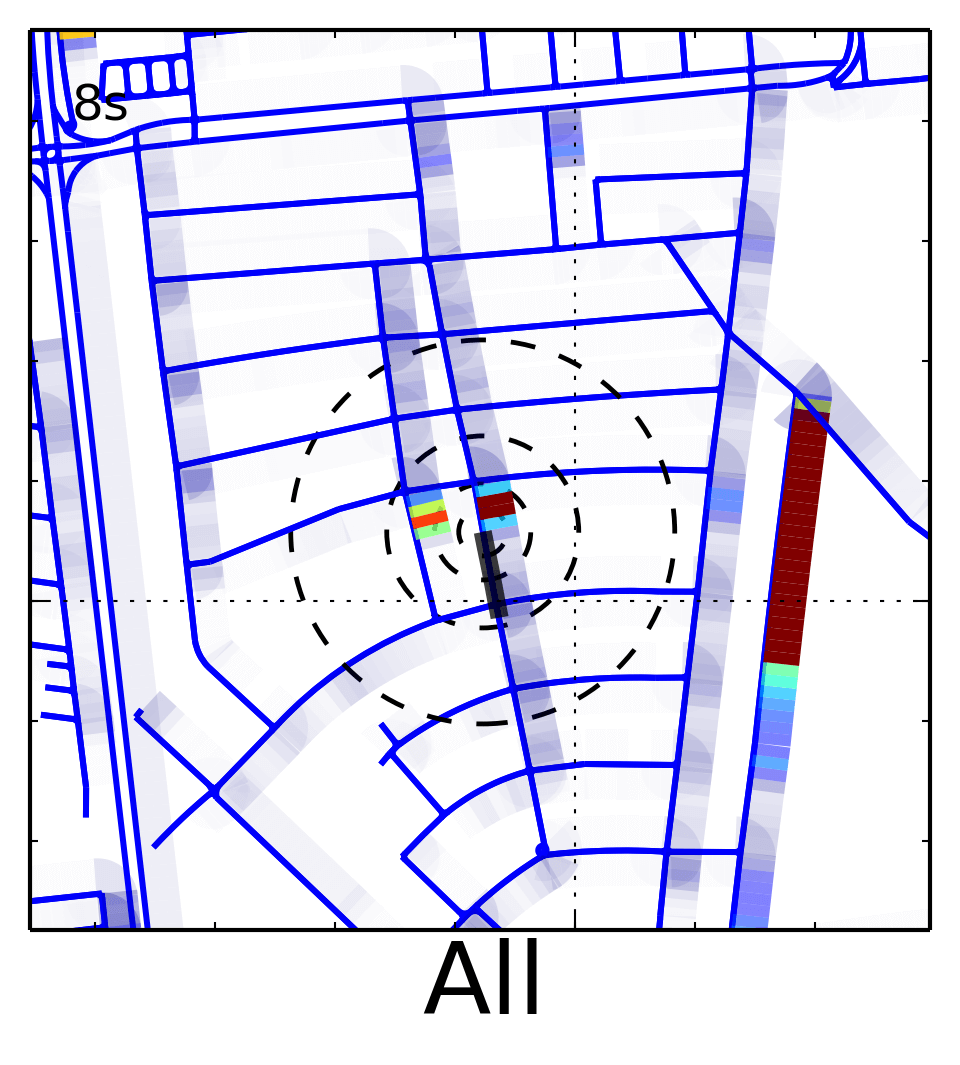}\\
\raisebox{26px}{\rotatebox{90}{$t = 18$s}}
&\includegraphics[width=0.16\linewidth,  trim={0 0.6cm 0 0}]{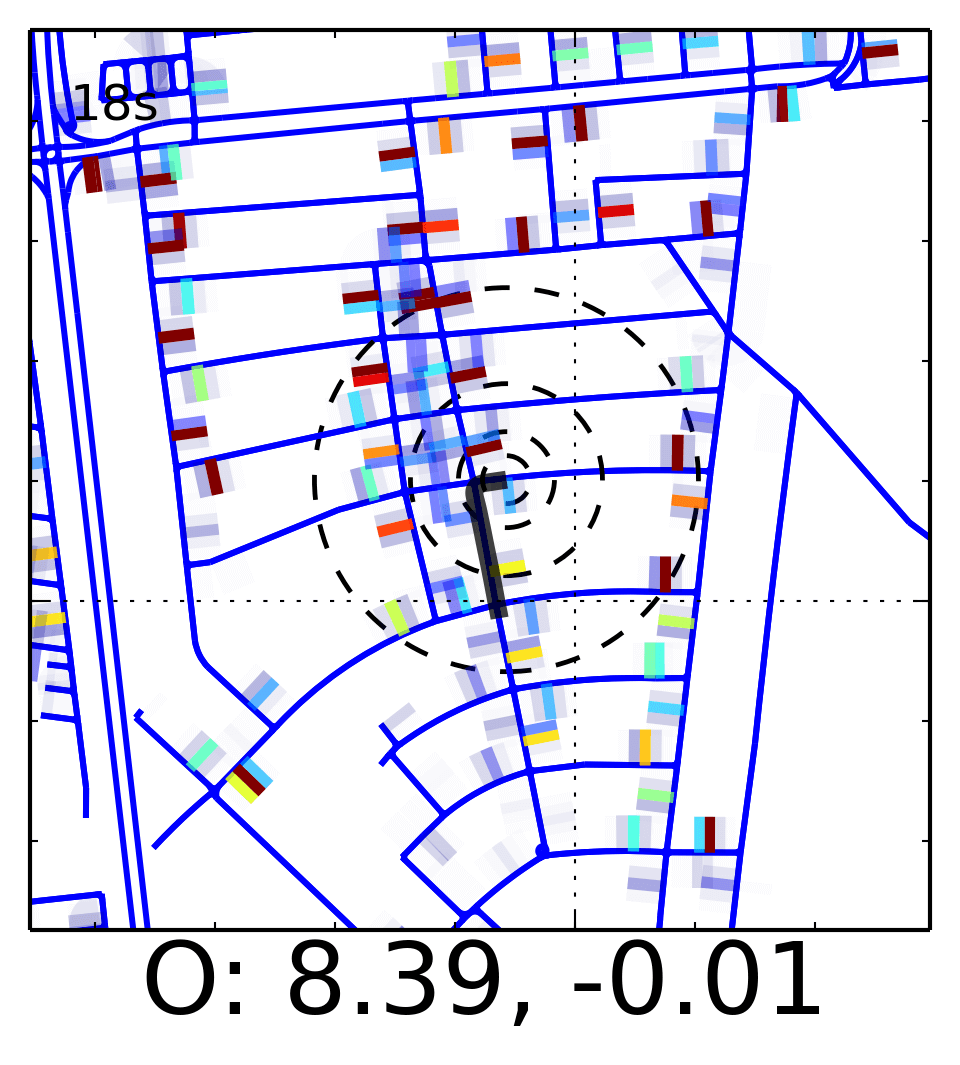}
&\includegraphics[width=0.16\linewidth,  trim={0 0.6cm 0 0}]{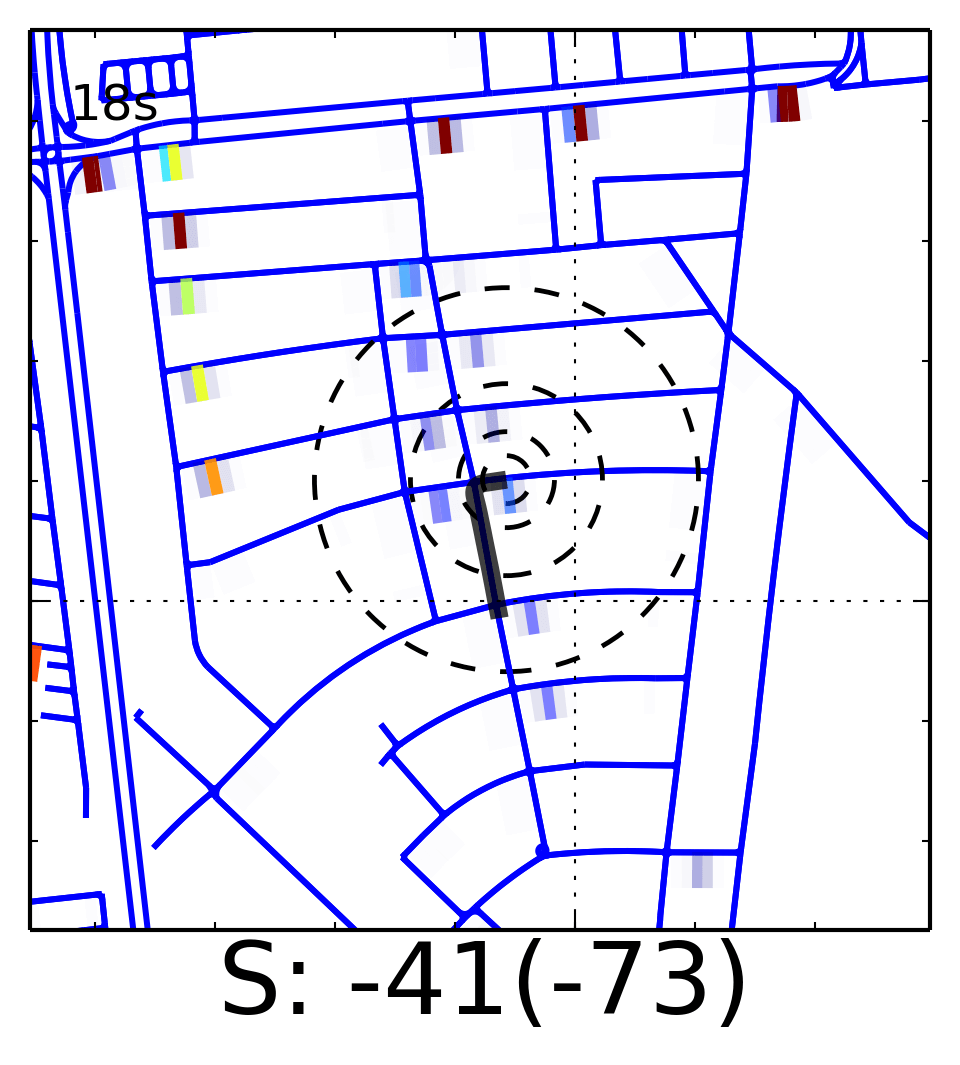}
&\includegraphics[width=0.16\linewidth,  trim={0 0.6cm 0 0}]{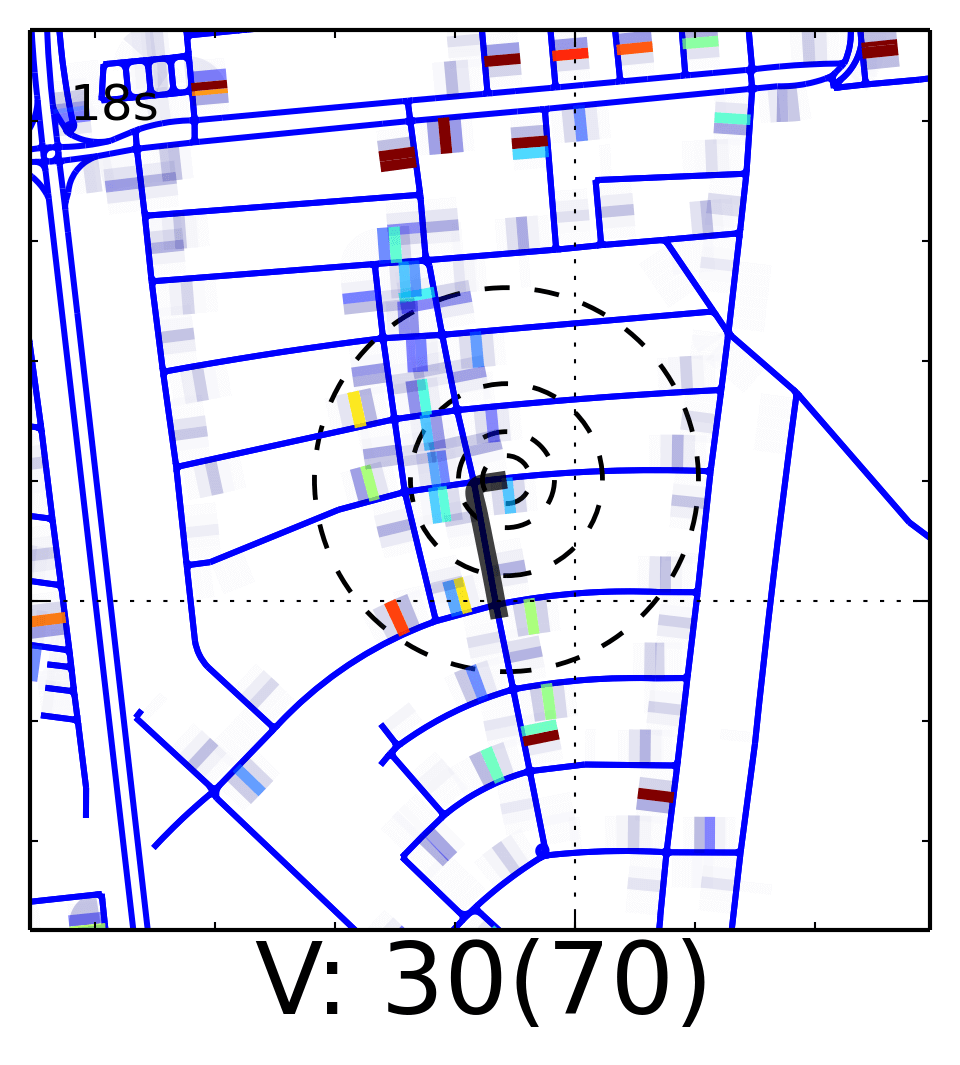}
&\includegraphics[width=0.16\linewidth,  trim={0 0.6cm 0 0}]{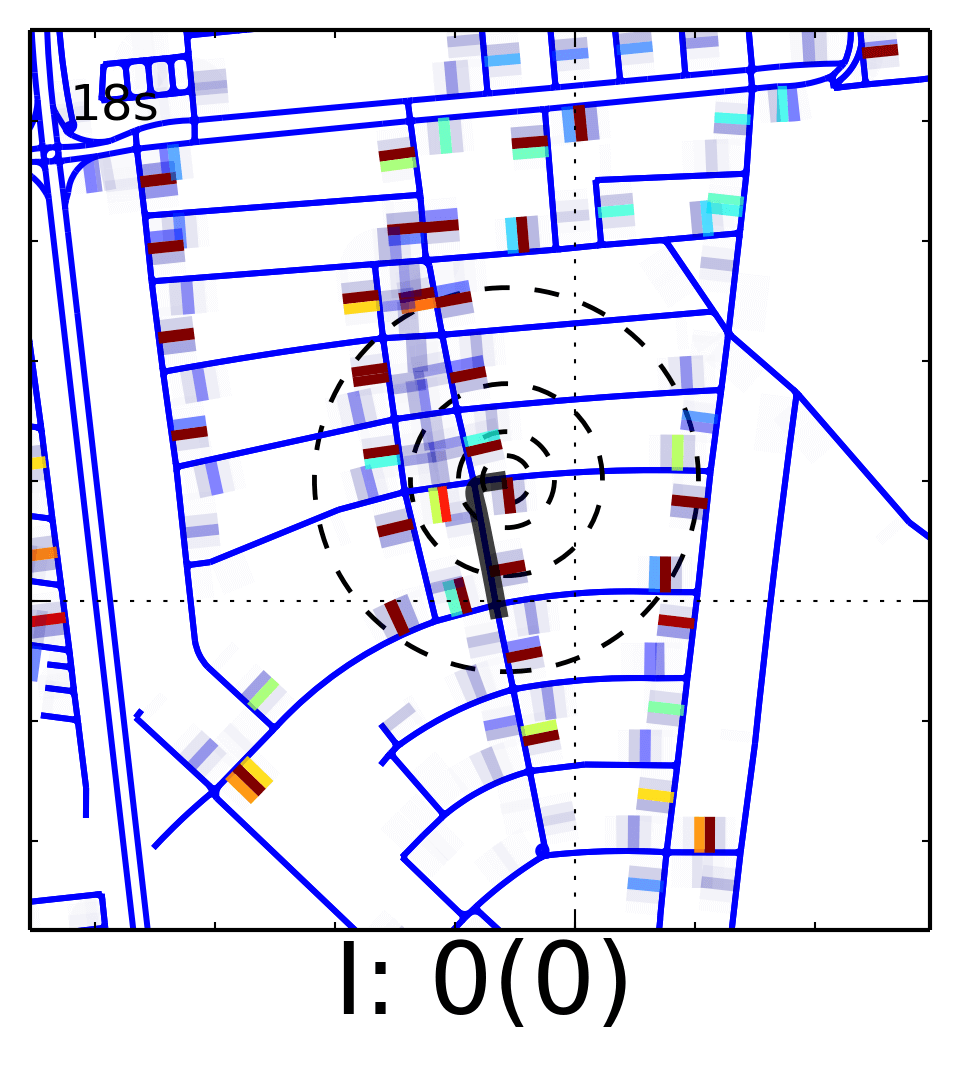}
&\includegraphics[width=0.16\linewidth,  trim={0 0.6cm 0 0}]{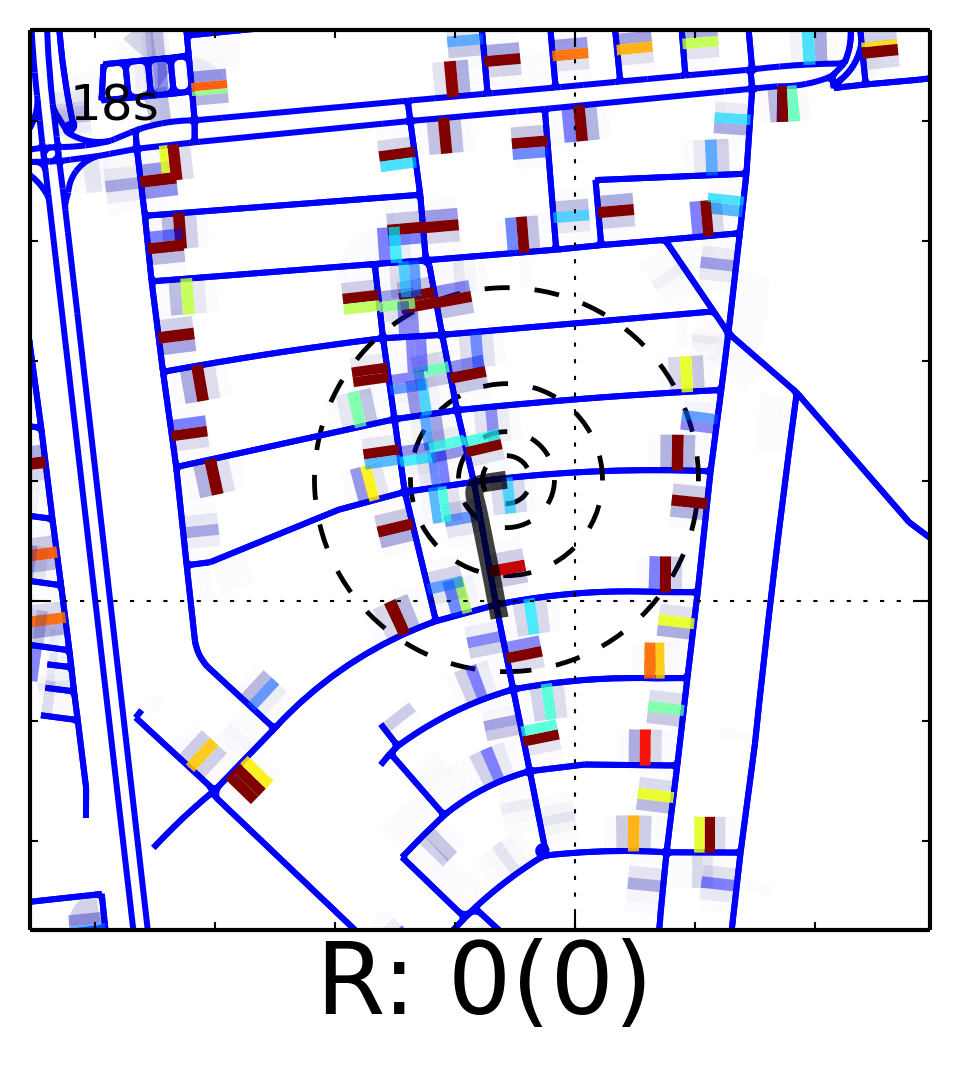}
&\includegraphics[width=0.16\linewidth,  trim={0 0.6cm 0 0}]{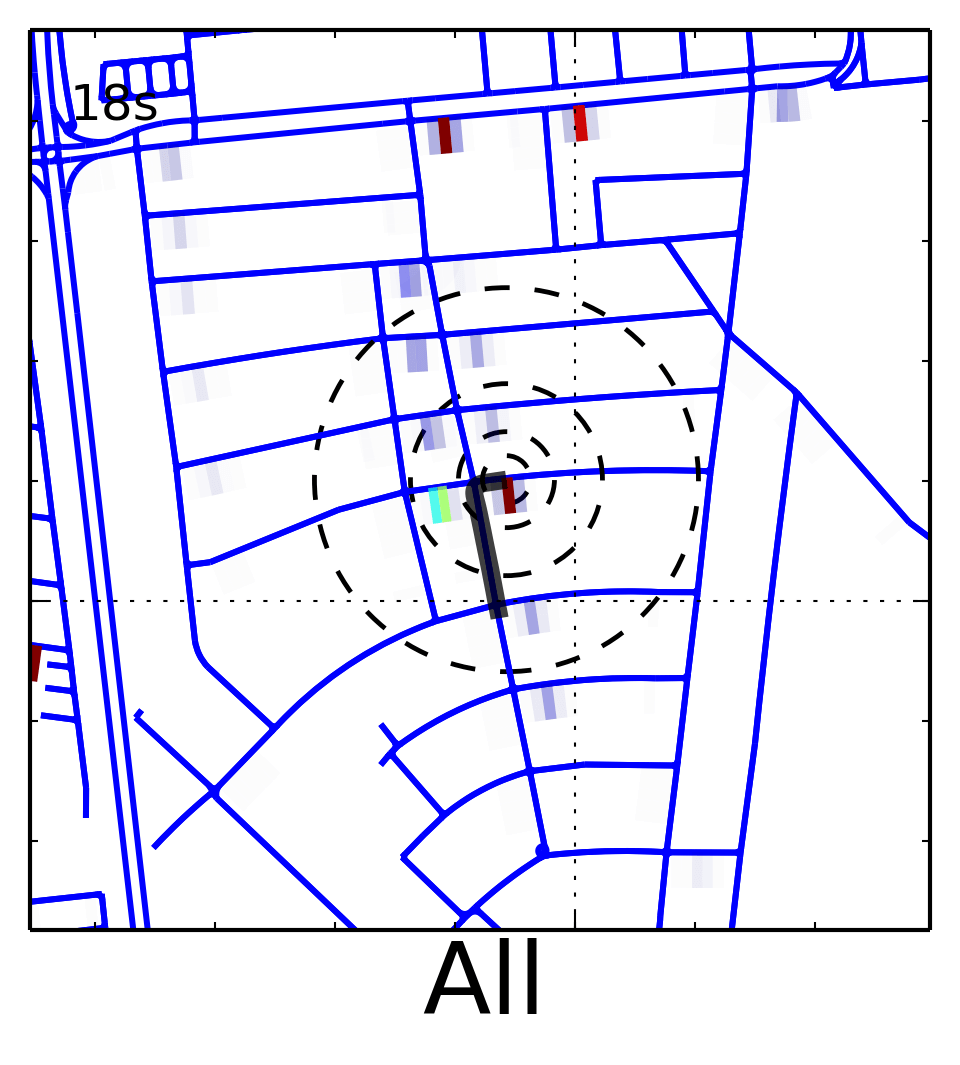}\\
\raisebox{26px}{\rotatebox{90}{$t = 26$s}}
&\includegraphics[width=0.16\linewidth,  trim={0 0.6cm 0 0}]{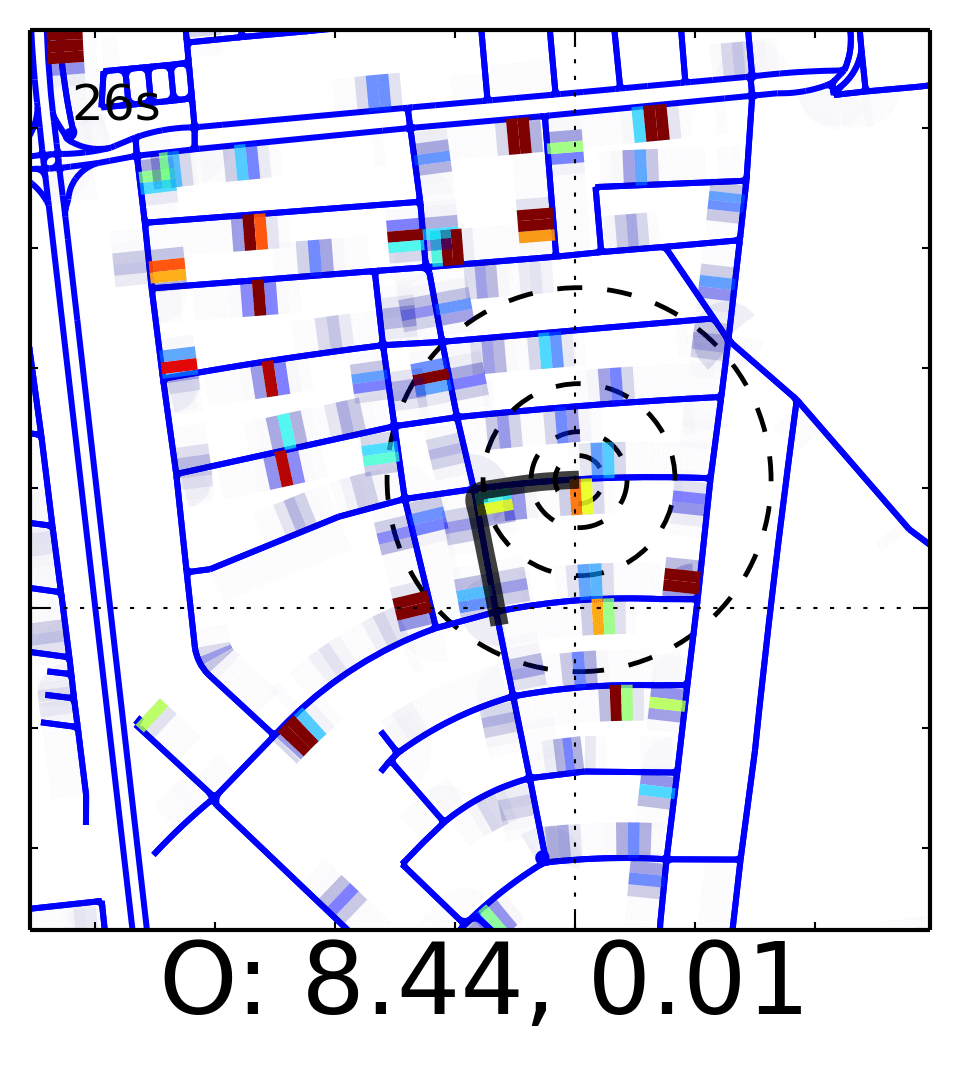}
&\includegraphics[width=0.16\linewidth,  trim={0 0.6cm 0 0}]{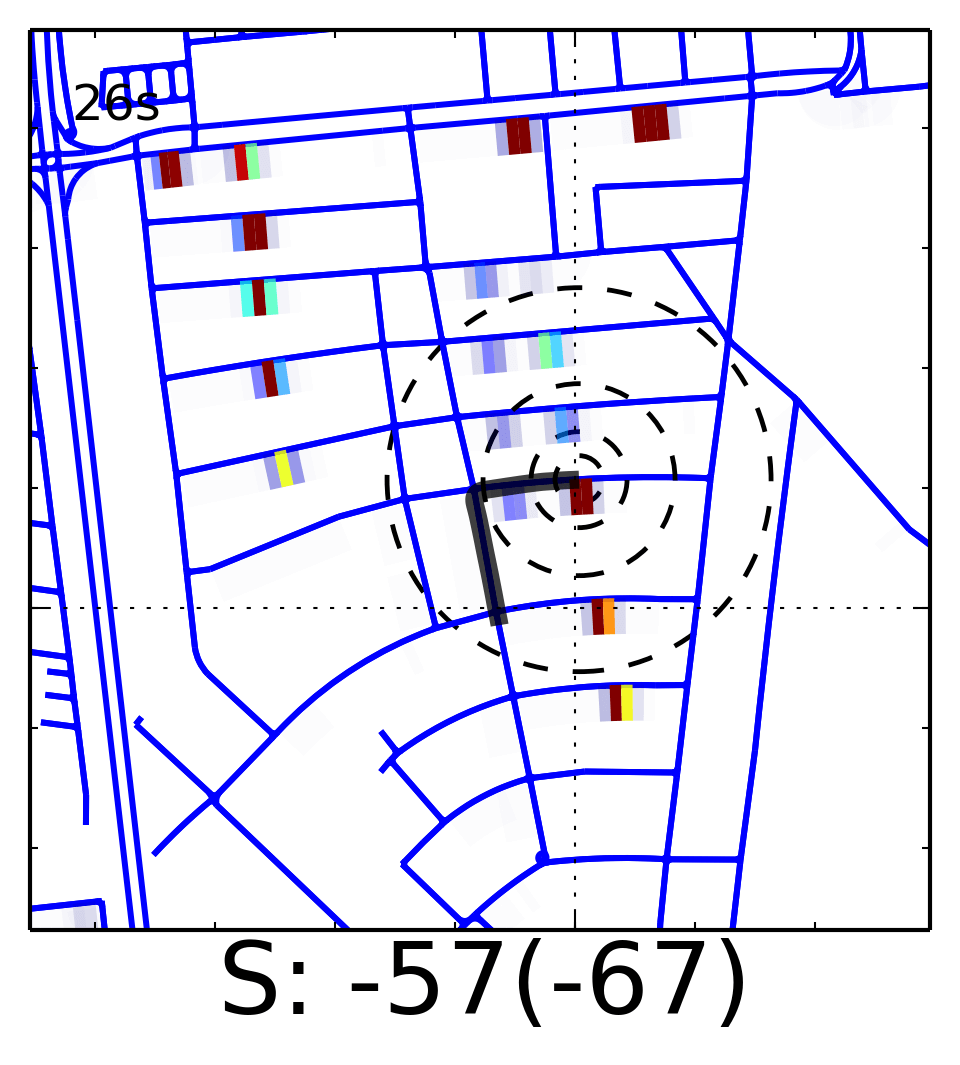}
&\includegraphics[width=0.16\linewidth,  trim={0 0.6cm 0 0}]{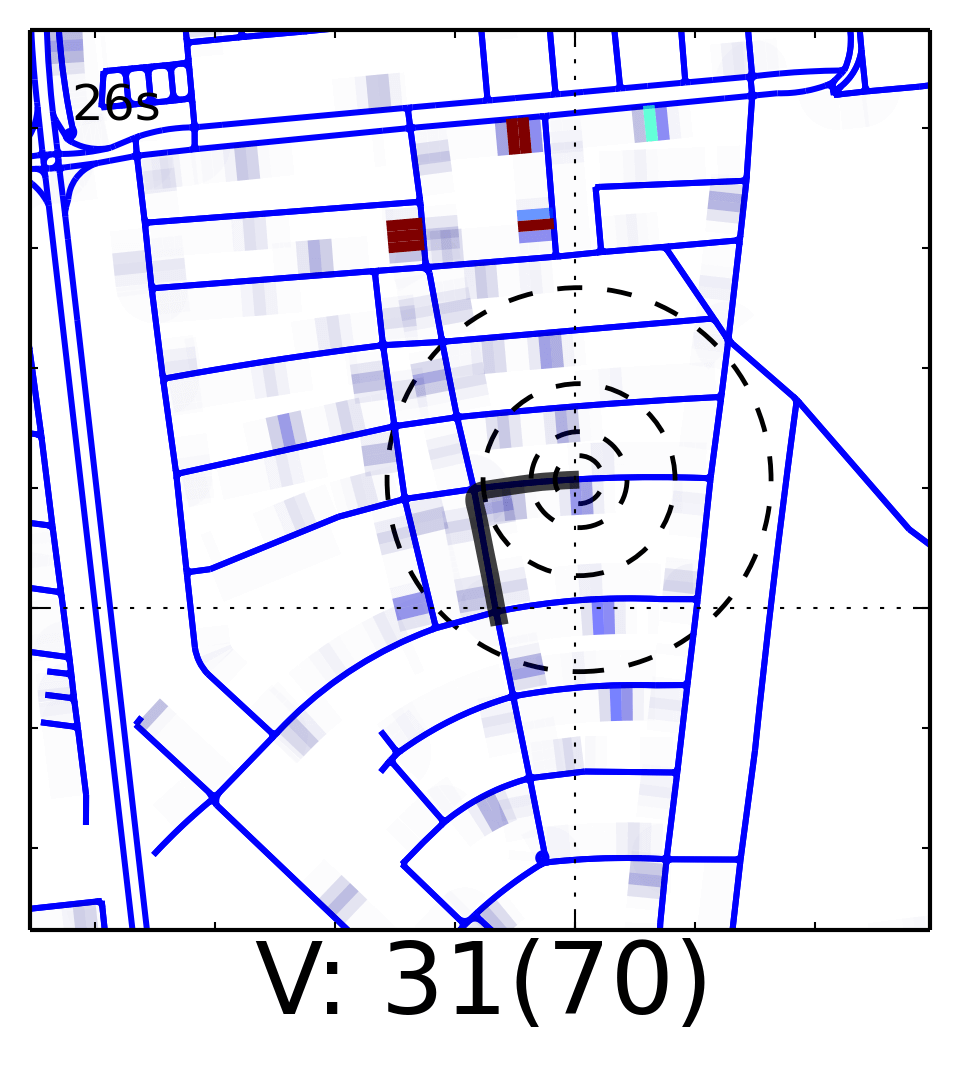}
&\includegraphics[width=0.16\linewidth,  trim={0 0.6cm 0 0}]{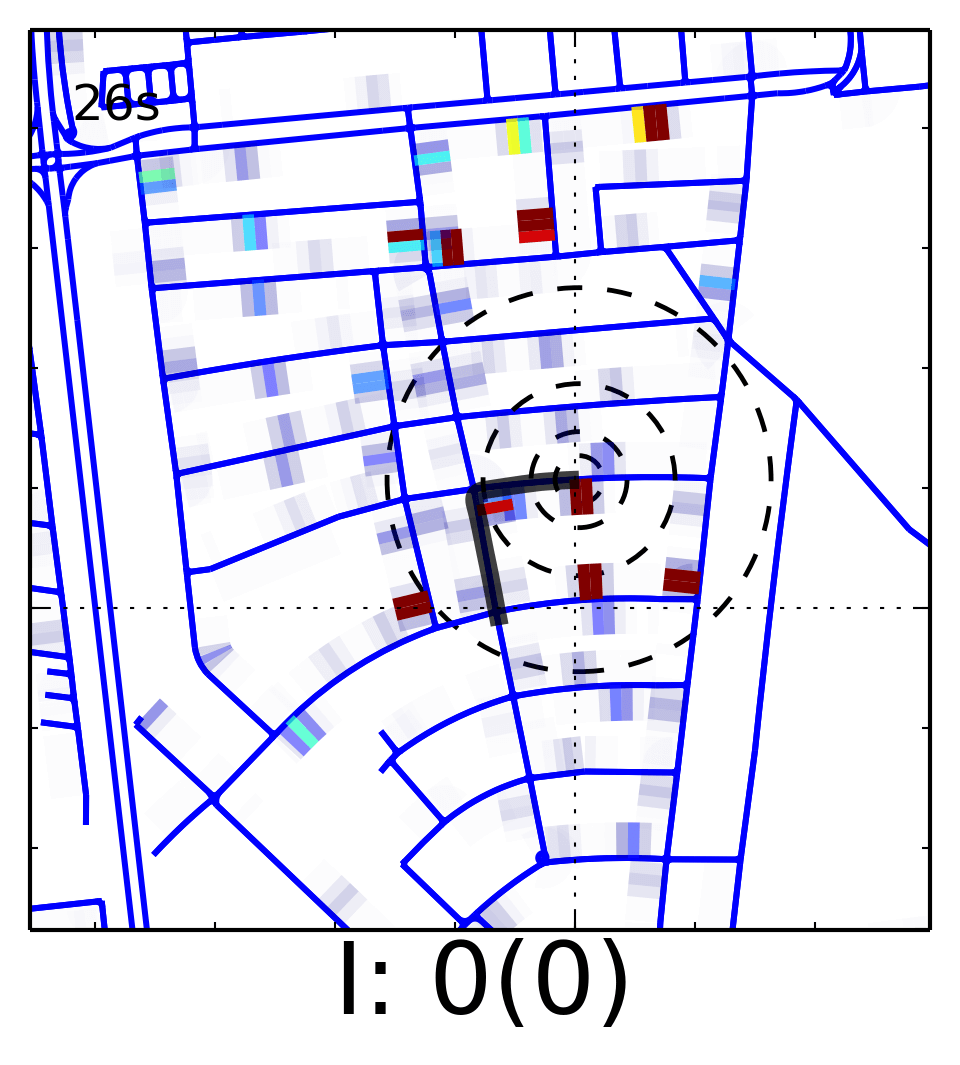}
&\includegraphics[width=0.16\linewidth,  trim={0 0.6cm 0 0}]{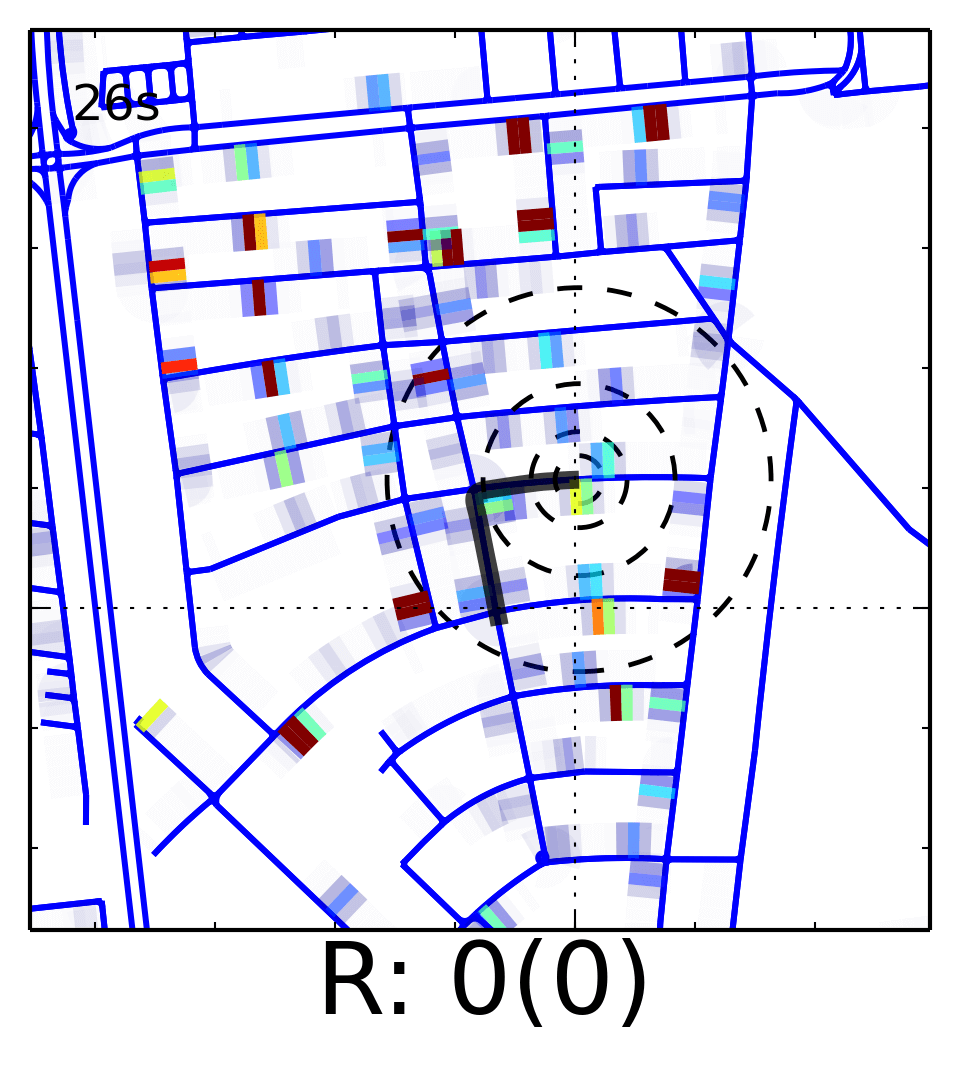}
&\includegraphics[width=0.16\linewidth,  trim={0 0.6cm 0 0}]{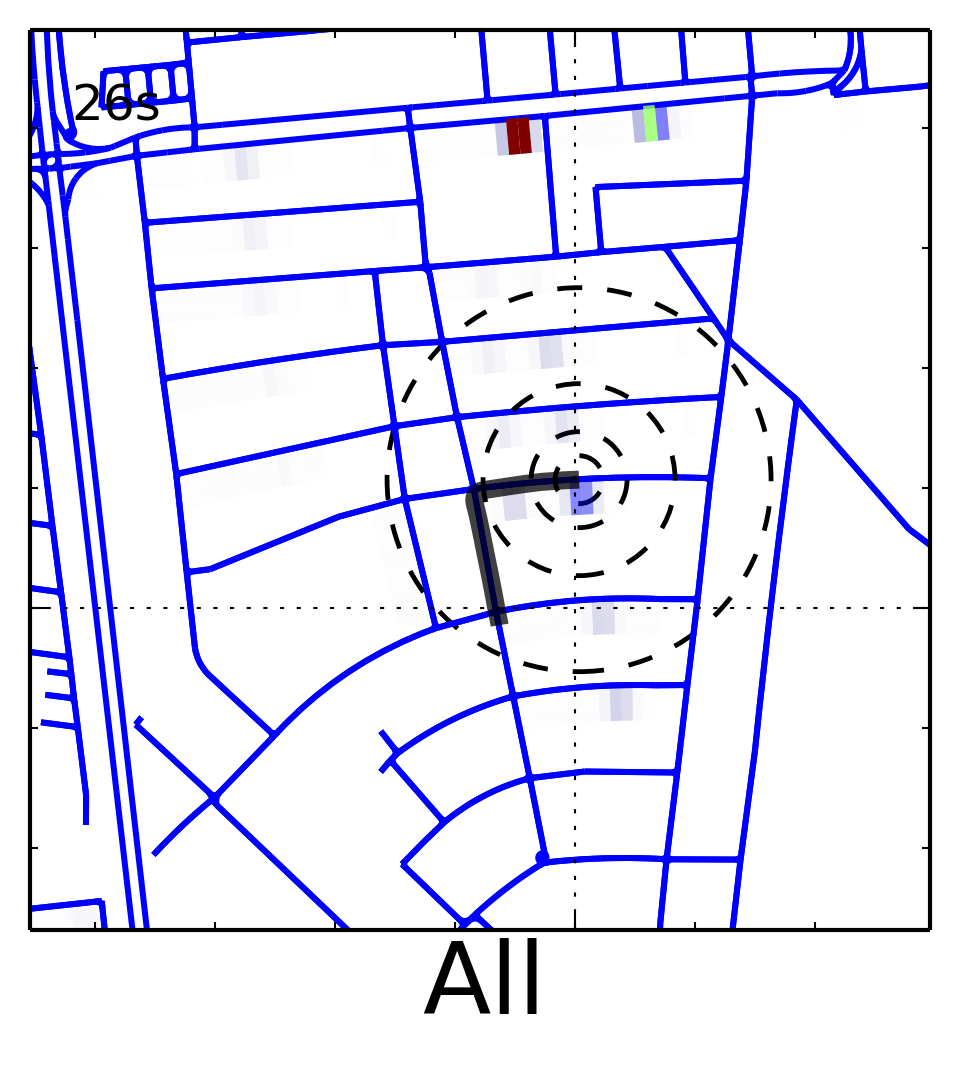}\\
\end{tabular}
\vspace{-0.3cm}
\caption{{\bf Contribution of each semantics:} Results using different semantics at several time steps. The different semantics help reduce the uncertainty in the map. The observations and the ground truth (in parenthesis) are shown below each image.}
\vspace{-4mm}
\label{fig:semantic-analysis}
\end{figure*}

To visualize how individual cues contribute to the localization task, we show  frames from  sequence 05 in \figref{fig:semantic-analysis}.  Sun direction provides the strongest initial cue, as it is able to fairly quickly eliminate a large number of roads whose directions are inconsistent with the heading implied by the sun.  Speed limit information is most helpful when a vehicle is traveling at high speeds, enabling many low-speed roads to be pruned.  However, even when traveling at slower speeds,  it can be helpful as slower speeds are less likely on highways.  Thus the portion of highway on the left  has lower likelihood with speed information than with odometry alone.  Intersection cues also help pruning the space.  For instance, at $t=26$s the vehicle is not near an intersection and thus one can observe that modes near intersections are suppressed.  Finally, the road type classifier by its nature is only useful in differentiating highways from other roads. For instance, at $t=8$s the highway has been pruned.

\begin{figure}[tb]
\vspace{-0.7cm}
\begin{subfigure}{0.2\textwidth}
\includegraphics[width=0.8\linewidth]{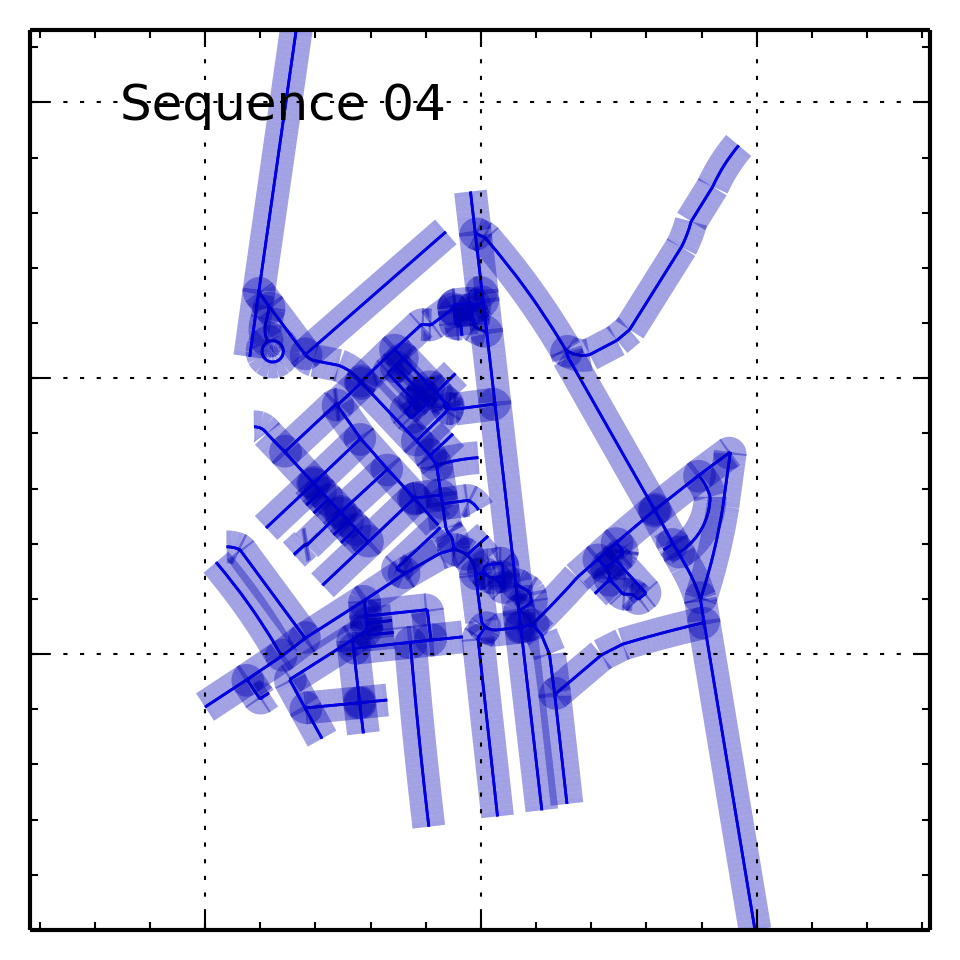}
\end{subfigure}
\hspace{-7 mm}\hfill
\begin{subfigure}{0.4\textwidth}
\includegraphics[width=0.25\linewidth]{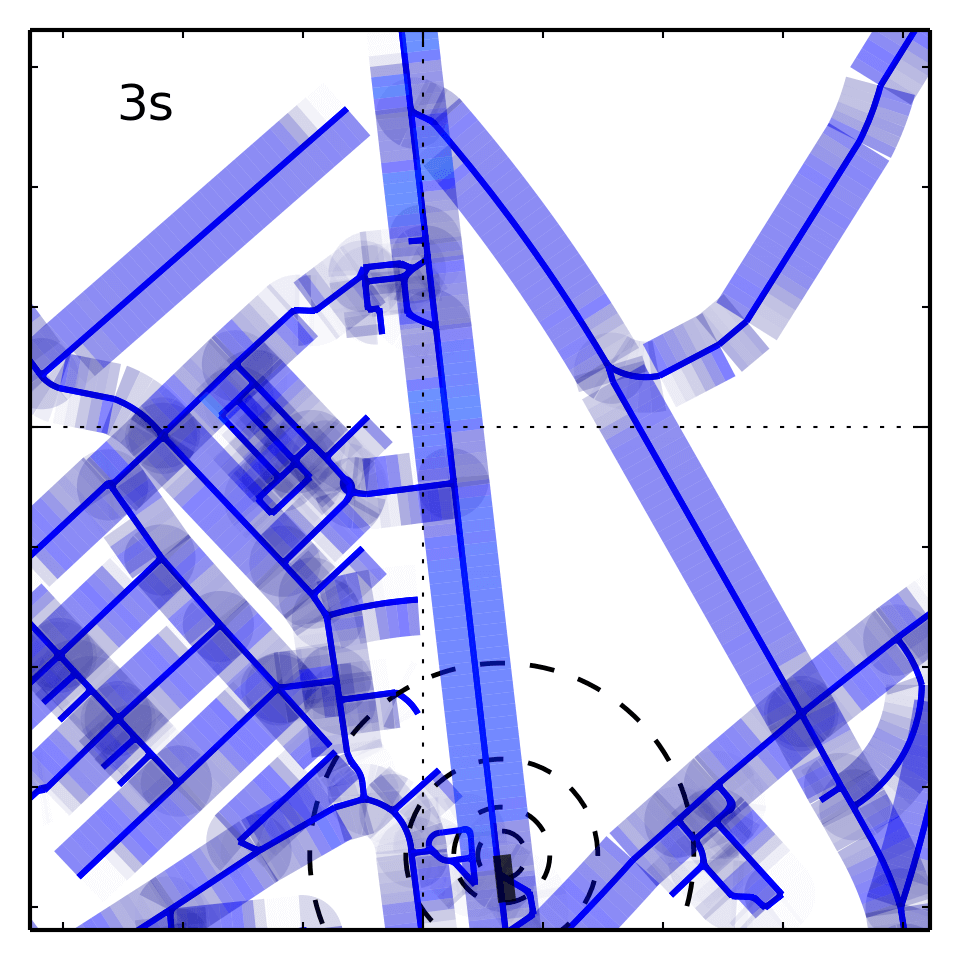}
\includegraphics[width=0.25\linewidth]{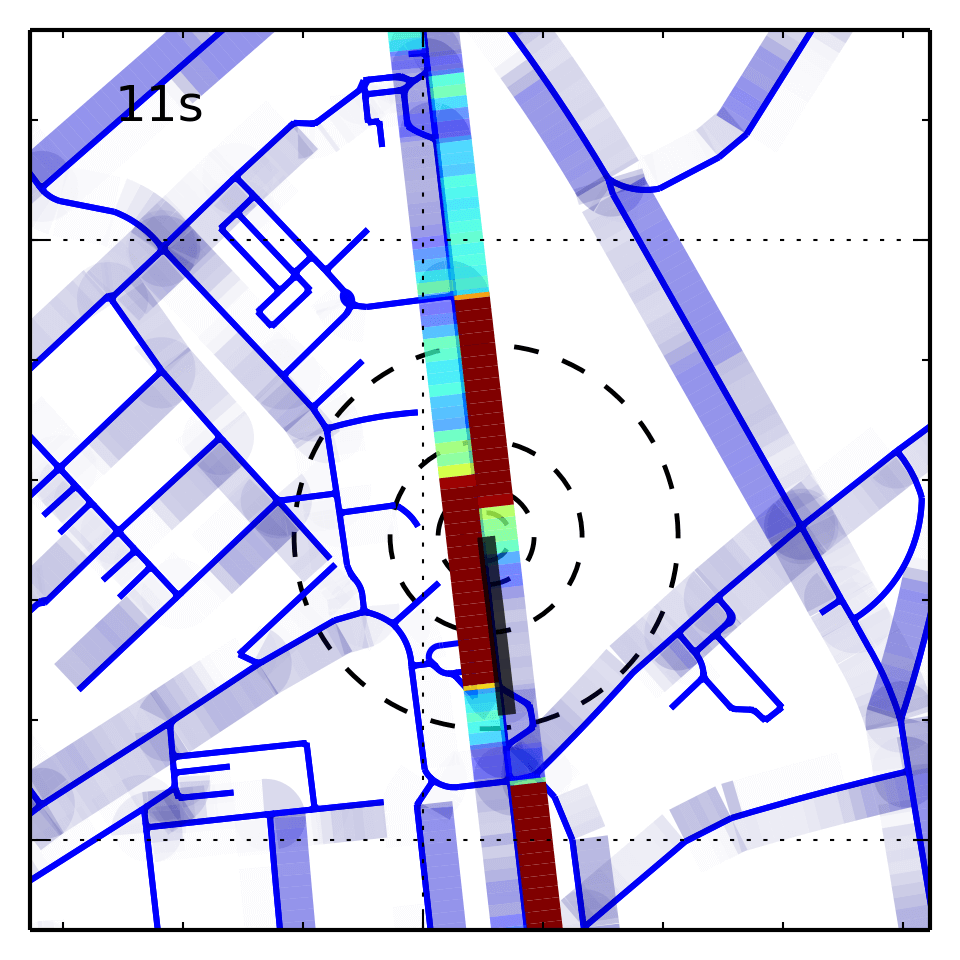}
\includegraphics[width=0.25\linewidth]{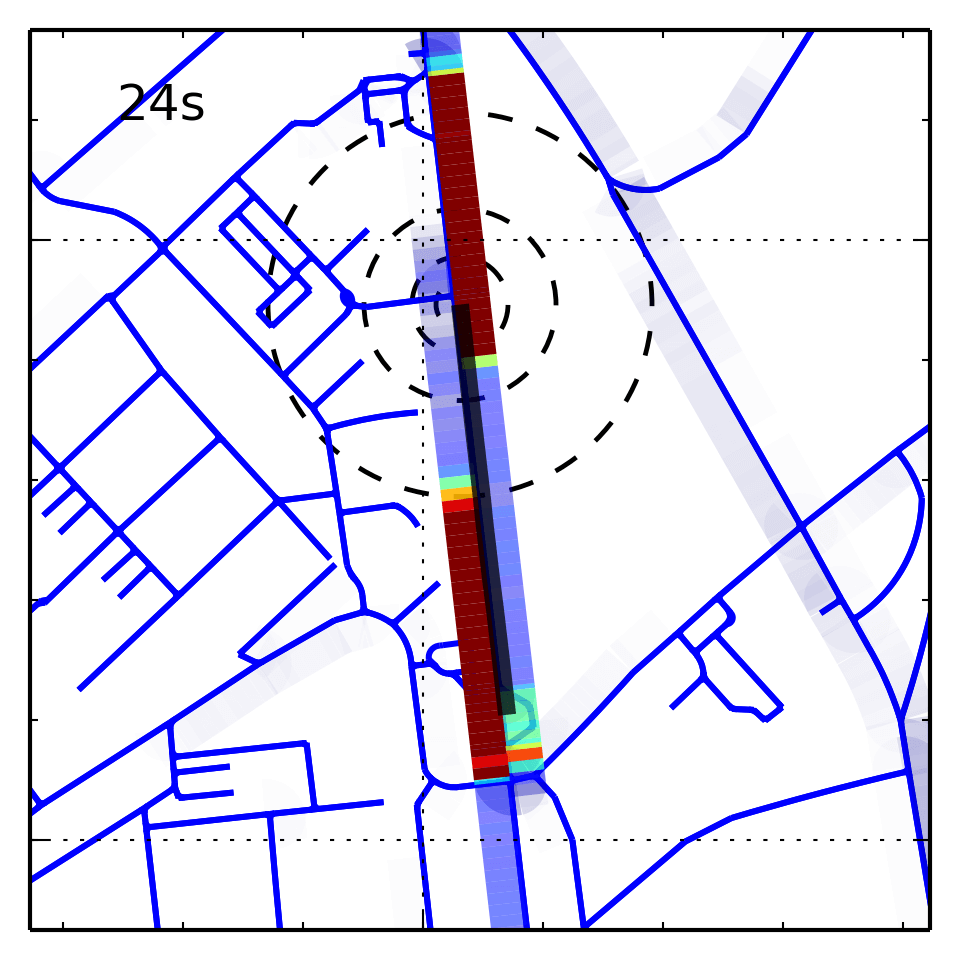}\\
\includegraphics[width=0.25\linewidth]{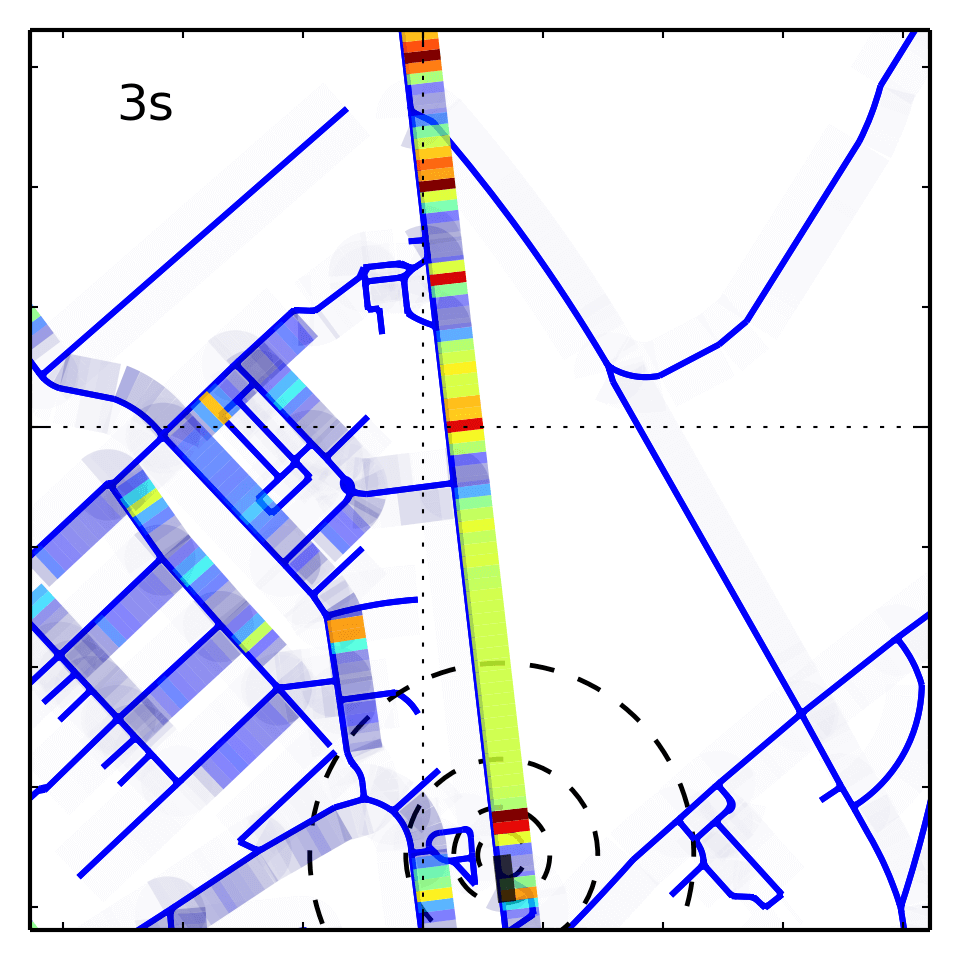}
\includegraphics[width=0.25\linewidth]{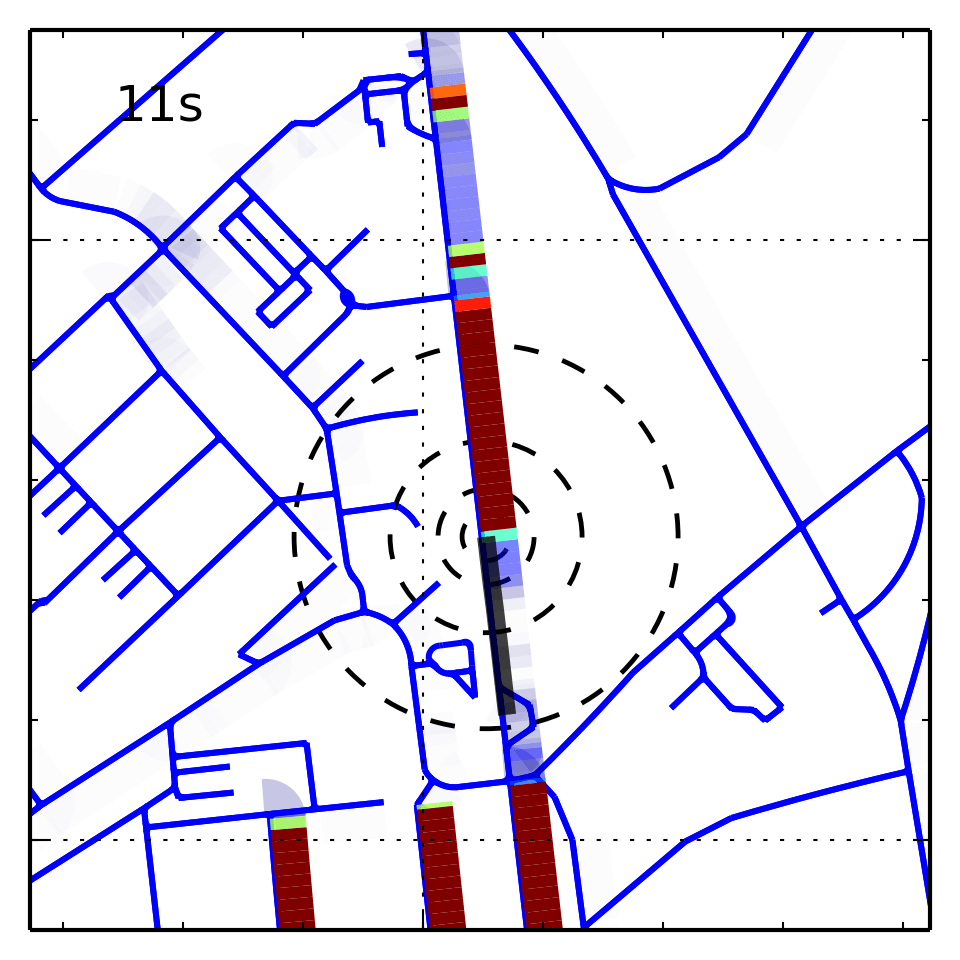}
\includegraphics[width=0.25\linewidth]{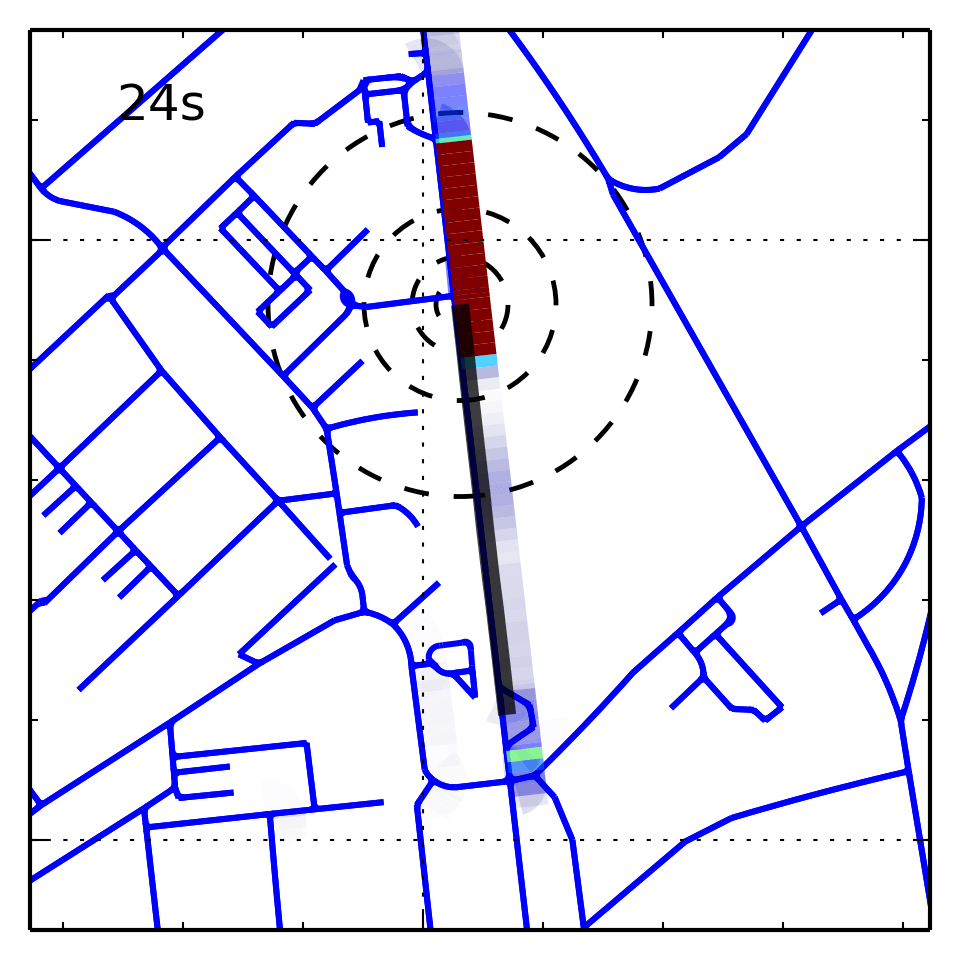}\\
\end{subfigure}\\
\vspace{-6mm}
\caption{{\bf Failure case:} The first row shows the results of~\cite{brubaker2013lost}, while the second row shows ours with all semantics. Although our model still fails to localize, we have reduced considerably the uncertainty. Just a few more seconds of driving, we would be able to localize by pruning the faint secondary mode using the intersection information.}
\vspace{-4mm}
\label{fig:failure}
\end{figure}

While the added cues make localization more robust and efficient, there are still limitations, particularly for extremely short or ambiguous sequences.  \figref{fig:failure} shows a sequence which still fails to localize, even with the added cues.  We note that when compared to the odometry only case (top) the uncertainty has been further reduced (bottom), suggesting that with just a few more seconds of driving we would be able to  localize by pruning out the faint secondary mode.  In contrast, with only odometry, there still remains the ambiguity of the overall direction of travel.

%% file: conc.tex
\section{Conclusions and Future Work}

We have presented an effective affordable approach to self-localization, which exploits freely available maps as well as visual odometry and semantic cues such as the sun direction, presence/absence of an intersection, road type and speed limits. Towards this goal, we exploited deep learning to directly estimate the semantics from monocular images. Interestingly, our  Sun-CNN automatically learns coarse shading/shadow detectors. 
 We have also shown how to automatically generate high quality labels without human intervention for all semantic tasks. We demonstrated the effectiveness of our localization approach in the challenging KITTI dataset and showed that we can localize faster and using less computation than \cite{brubaker2013lost}.
In the future, we plan to exploit more semantic cues like traffic signs, trees, buildings, and road width. Although these data are either sparsely annotated or mostly incorrect, we hope the OSM maps  become more complete and noise-free.

%% file: top.bbl
\begin{thebibliography}{10}\itemsep=-1pt

\bibitem{Baatz12}
G.~Baatz, K.~K{\"o}ser, D.~Chen, R.~Grzeszczuk, and M.~Pollefeys.
\newblock {Leveraging 3D City Models for Rotation Invariant Place-of-Interest
  Recognition}.
\newblock {\em IJCV}, 2012.

\bibitem{Bansal14}
M.~Bansal and K.~Daniilidis.
\newblock Geometric urban geo-localization.
\newblock In {\em CVPR}, 2014.

\bibitem{Bonnabel2011}
S.~Bonnabel and E.~Sala{\"u}n.
\newblock Design and prototyping of a low-cost vehicle localization system with
  guaranteed convergence properties.
\newblock {\em Control Engineering Practice}, 19(6):591--601, 2011.

\bibitem{brubaker2013lost}
M.~Brubaker, A.~Geiger, and R.~Urtasun.
\newblock Lost! leveraging the crowd for probabilistic visual
  self-localization.
\newblock In {\em CVPR}, pages 3057--3064. IEEE, 2013.

\bibitem{castaldo_vss15}
F.~Castaldo, A.~R. Zamir, R.~Angst, F.~Palmieri, and S.~Savarese.
\newblock Semantic cross-view matching.
\newblock In {\em Vision from Satellite to Street - The Second International
  Workshop in Conjunction with ICCV 2015}, 2015.

\bibitem{Dellaert99}
F.~Dellaert, W.~Burgard, D.~Fox, and S.~Thrun.
\newblock Using the condensation algorithm for robust, vision-based mobile
  robot localization.
\newblock {\em CVPR}, 1999.

\bibitem{deng2009imagenet}
J.~Deng, W.~Dong, R.~Socher, L.-J. Li, K.~Li, and L.~Fei-Fei.
\newblock Imagenet: A large-scale hierarchical image database.
\newblock In {\em CVPR}, pages 248--255. IEEE, 2009.

\bibitem{Dewri13}
R.~Dewri, P.~Annadata, W.~Eltarjaman, and R.~Thurimella.
\newblock Inferring trip destinations from driving habits data.
\newblock In {\em WPES}, 2013.

\bibitem{ElNajjar2005}
M.~E. {El Najjar} and P.~Bonnifait.
\newblock A road-matching method for precise vehicle localization using belief
  theory and kalman filtering.
\newblock {\em Autonomous Robots}, 19(2):173--191, 2005.

\bibitem{floros13}
G.~Floros, B.~van~der Zander, and B.~Leibe.
\newblock {O}pen{S}treet{SLAM}: {G}lobal vehicle localization using
  {O}pen{S}treet{M}aps.
\newblock In {\em ICRA}, 2013.

\bibitem{Fouque2012}
C.~Fouque and P.~Bonnifait.
\newblock Matching raw gps measurements on a navigable map without computing a
  global position.
\newblock {\em IEEE Trans. on Intelligent Transportation Systems},
  13(2):887--898, 2012.

\bibitem{Fox99}
D.~Fox, W.~Burgard, F.~Dellaert, and S.~Thrun.
\newblock Monte carlo localization: Efficient position estimation for mobile
  robots.
\newblock In {\em AAAI}, 1999.

\bibitem{gastwirth1972estimation}
J.~L. Gastwirth.
\newblock The estimation of the lorenz curve and gini index.
\newblock {\em The Review of Economics and Statistics}, pages 306--316, 1972.

\bibitem{Geiger2012CVPR}
A.~Geiger, P.~Lenz, and R.~Urtasun.
\newblock Are we ready for autonomous driving? the kitti vision benchmark
  suite.
\newblock In {\em CVPR}, 2012.

\bibitem{geiger2011stereoscan}
A.~Geiger, J.~Ziegler, and C.~Stiller.
\newblock Stereoscan: Dense 3d reconstruction in real-time.
\newblock In {\em Intelligent Vehicles Symposium}, pages 963--968. IEEE, 2011.

\bibitem{Guivant2007}
J.~Guivant and R.~Katz.
\newblock Global urban localization based on road maps.
\newblock In {\em IROS}, pages 1079--1084, 2007.

\bibitem{Gutmann98}
J.-S. Gutmann, W.~Burgard, D.~Fox, and K.~Konolige.
\newblock An experimental comparison of localization methods.
\newblock In {\em ICIRS}, 1998.

\bibitem{Hays08}
J.~Hays and A.~A. Efros.
\newblock im2gps: estimating geographic information from a single image.
\newblock In {\em CVPR}, 2008.

\bibitem{JunejoECCV2008}
I.~N. Junejo and H.~Foroosh.
\newblock Estimating geo-temporal location of stationary cameras using shadow
  trajectories.
\newblock In {\em Computer Vision--ECCV 2008}, pages 318--331. Springer, 2008.

\bibitem{krizhevsky2012imagenet}
A.~Krizhevsky, I.~Sutskever, and G.~E. Hinton.
\newblock Imagenet classification with deep convolutional neural networks.
\newblock In {\em NIPS}, pages 1097--1105, 2012.

\bibitem{lalonde2009estimating}
J.-F. Lalonde, A.~A. Efros, and S.~G. Narasimhan.
\newblock Estimating natural illumination from a single outdoor image.
\newblock In {\em ICCV}, 2009.

\bibitem{LambertJFR2012}
A.~Lambert, P.~Furgale, T.~Barfoot, and J.~Enright.
\newblock Field testing of visual odometry aided by a sun sensor and
  inclinometer.
\newblock {\em Journal of Field Robotics}, 29:426 -- 444, 2012.

\bibitem{Li06}
F.~Li and J.~Kosecka.
\newblock Probabilistic location recognition using reduced feature set.
\newblock In {\em ICRA}, 2006.

\bibitem{Li12}
Y.~Li, N.~Snavely, D.~Huttenlocher, and P.~Fua.
\newblock Worldwide pose estimation using 3d point clouds.
\newblock In {\em ECCV}, 2012.

\bibitem{lin2015learning}
T.-Y. Lin, Y.~Cui, S.~Belongie, J.~Hays, and C.~Tech.
\newblock Learning deep representations for ground-to-aerial geolocalization.
\newblock In {\em CVPR}, pages 5007--5015, 2015.

\bibitem{linegar2015work}
C.~Linegar, W.~Churchill, and P.~Newman.
\newblock Work smart, not hard: Recalling relevant experiences for vast-scale
  but time-constrained localisation.
\newblock In {\em Robotics and Automation (ICRA), 2015 IEEE International
  Conference on}, pages 90--97. IEEE, 2015.

\bibitem{ApartmentsCVPR15}
C.~Liu, A.~Schwing, K.~Kundu, R.~Urtasun, and S.~Fidler.
\newblock Rent3d: Floor-plan priors for monocular layout estimation.
\newblock In {\em CVPR}, 2015.

\bibitem{Brualla14}
R.~Martin-Brualla, Y.~He, B.~C. Russell, and S.~M. Seitz.
\newblock {The 3D Jigsaw Puzzle: Mapping Large Indoor Spaces}.
\newblock In {\em ECCV}, 2014.

\bibitem{nyc3dcar}
K.~Matzen and N.~Snavely.
\newblock Nyc3dcars: A dataset of 3d vehicles in geographic context.
\newblock In {\em ICCV}, 2013.

\bibitem{moosmann2013joint}
F.~Moosmann and C.~Stiller.
\newblock Joint self-localization and tracking of generic objects in 3d range
  data.
\newblock In {\em Robotics and Automation (ICRA), 2013 IEEE International
  Conference on}, pages 1146--1152. IEEE, 2013.

\bibitem{nelson2015dusk}
P.~Nelson, W.~Churchill, I.~Posner, and P.~Newman.
\newblock From dusk till dawn: Localisation at night using artificial light
  sources.
\newblock In {\em Robotics and Automation (ICRA), 2015 IEEE International
  Conference on}, pages 5245--5252. IEEE, 2015.

\bibitem{Oh04}
S.~M. Oh, S.~Tariq, B.~N. Walker, and F.~Dellaert.
\newblock Map-based priors for localization.
\newblock In {\em ICIRS}, 2004.

\bibitem{solar}
I.~Reda and A.~Andreas.
\newblock Solar position algorithm for solar radiation applications.
\newblock {\em Solar energy}, 76(5):577--589, 2004.

\bibitem{Sattler11}
T.~Sattler, B.~Leibe, and L.~Kobbelt.
\newblock Fast image-based localization using direct 2d-to-3d matching.
\newblock In {\em ICCV}, 2011.

\bibitem{Schindler07}
G.~Schindler, M.~Brown, and R.~Szeliski.
\newblock City-scale location recognition.
\newblock In {\em CVPR}, 2007.

\bibitem{szegedy2014going}
C.~Szegedy, W.~Liu, Y.~Jia, P.~Sermanet, S.~Reed, D.~Anguelov, D.~Erhan,
  V.~Vanhoucke, and A.~Rabinovich.
\newblock Going deeper with convolutions.
\newblock {\em arXiv preprint arXiv:1409.4842}, 2014.

\bibitem{van2008visualizing}
L.~Van~der Maaten and G.~Hinton.
\newblock Visualizing data using t-sne.
\newblock {\em Journal of Machine Learning Research}, 9(2579-2605):85, 2008.

\bibitem{WangCVPR15}
S.~Wang, S.~Fidler, and R.~Urtasun.
\newblock Holistic 3d scene understanding from a single geo-tagged image.
\newblock In {\em CVPR}, 2015.

\bibitem{wolcott2014visual}
R.~W. Wolcott and R.~M. Eustice.
\newblock Visual localization within lidar maps for automated urban driving.
\newblock In {\em Intelligent Robots and Systems (IROS 2014), 2014 IEEE/RSJ
  International Conference on}, pages 176--183. IEEE, 2014.

\bibitem{WorkmanECCV2014}
S.~Workman, R.~P. Mihail, and N.~Jacobs.
\newblock A pot of gold: Rainbows as a calibration cue.
\newblock In {\em ECCV}, pages 820--835. Springer, 2014.

\bibitem{workman2015wide}
S.~Workman, R.~Souvenir, and N.~Jacobs.
\newblock Wide-area image geolocalization with aerial reference imagery.
\newblock {\em arXiv preprint arXiv:1510.03743}, 2015.

\bibitem{WuCVIU2010}
L.~Wu, X.~Cao, and H.~Foroosh.
\newblock Camera calibration and geo-location estimation from two shadow
  trajectories.
\newblock {\em Computer Vision and Image Understanding}, 114(8):915--927, 2010.

\bibitem{Zhang06}
W.~Zhang and J.~Kosecka.
\newblock Image based localization in urban environments.
\newblock In {\em 3DPVT}, 2006.

\bibitem{zhou2015object}
B.~Zhou, A.~Khosla, A.~Lapedriza, A.~Oliva, and A.~Torralba.
\newblock Object detectors emerge in deep scene cnns.
\newblock In {\em ICLR}, 2015.

\bibitem{zhou2014learning}
B.~Zhou, A.~Lapedriza, J.~Xiao, A.~Torralba, and A.~Oliva.
\newblock Learning deep features for scene recognition using places database.
\newblock In {\em NIPS}, pages 487--495, 2014.

\end{thebibliography}
